\documentclass[runningheads]{llncs}

\usepackage[mobile]{eccv}

\usepackage{eccvabbrv}

\usepackage{graphicx}
\usepackage[export]{adjustbox}
\usepackage{booktabs}
\usepackage{makecell}
\usepackage{tikz}
\usepackage{pgfplots}
\pgfplotsset{compat=1.18}

\definecolor{teal}{RGB}{0,128,128}
\definecolor{softblue}{RGB}{120,160,220}

\usepackage[accsupp]{axessibility}  

\usepackage[utf8]{inputenc}
\usepackage[T1]{fontenc}      
\usepackage{appendix}       
\usepackage{lipsum}   
\usepackage{multirow}
\usepackage{wrapfig}

\usepackage[pagebackref,breaklinks,colorlinks,citecolor=eccvblue]{hyperref}
\hypersetup{
   colorlinks=true,
   linkcolor=teal,
   citecolor=teal,
   urlcolor=teal
}

\usepackage{orcidlink}
\usepackage{booktabs}
\usepackage[table,x11names,dvipsnames,table]{xcolor}

\definecolor{ForestGreen}{RGB}{34,139,34}
\definecolor{Orange}{RGB}{1.0, 0.55, 0.0}
\definecolor{Purple}{RGB}{0.58, 0.44, 0.86}
\definecolor{softblue}{rgb}{0.2, 0.4, 0.8} 
\definecolor{lightblue}{rgb}{0.6, 0.75, 0.95}
\definecolor{darkblue}{rgb}{0.05, 0.2, 0.5}

\definecolor{coral}{RGB}{255, 127, 80} 
\definecolor{softcoral}{rgb}{1.00, 0.60, 0.50}  

\definecolor{teal}{rgb}{0.1, 0.6, 0.6} 
\definecolor{forestgreen}{rgb}{0.1, 0.6, 0.2} 

\definecolor{vibrantorange}{rgb}{1.0, 0.5, 0.0}

\definecolor{deepblue}{rgb}{0.0, 0.0, 0.8}

\definecolor{TableColor}{rgb}{0.58, 0.55, 0.51}

\definecolor{TableColor}{rgb}{0.92, 0.95, 0.99}  
\definecolor{TableColorGrey}{rgb}{0.88, 0.88, 0.88}  

\definecolor{DarkenedMagenta}{RGB}{225,0,100}

\newcommand{\Eq}[1]{\hyperref[#1]{Eq.~(\ref{#1})}}
\newcommand{\Equation}[1]{\hyperref[#1]{Equation~(\ref{#1})}}
\newcommand{\secref}[1]{\hyperref[#1]{Sec.~\ref{#1}}}

\newcommand{\Th}[1]{\textsc{#1}}

\newcommand{\red}[1]{{\textcolor{red}{#1}}}

\newcommand{\citeme}[1]{\red{[XX]}}
\newcommand{\refme}[1]{\red{(XX)}}

\makeatletter
\newcommand*\bdot{\mathpalette\bdot@{.7}}
\newcommand*\bdot@[2]{\mathbin{\vcenter{\hbox{\scalebox{#2}{$\m@th#1\bullet$}}}}}
\makeatother

\makeatletter
\DeclareRobustCommand\onedot{\futurelet\@let@token\@onedot}
\def\@onedot{\ifx\@let@token.\else.\null\fi\xspace}

\makeatother

\begin{document}

\title{Coevolving Representations in \\  Joint Image-Feature Diffusion} 

\titlerunning{Coevolving Representations}




\author{
  Theodoros Kouzelis$^{1,4}$\thanks{Corresponding author: \texttt{\url{theodoros.kouzelis@athenarc.gr}} \\ Code is available at \url{https://github.com/zelaki/CoReDi}} \and
  Spyros Gidaris$^{3}$ \and
  Nikos Komodakis$^{1,2,5}$ \\[6pt]
  {\small
    $^1$Archimedes, Athena RC \quad
    $^2$University of Crete \quad 
    $^3$valeo.ai \quad \\
    $^4$National Technical University of Athens \quad     
    $^5$IACM-Forth
  }
}

\authorrunning{T.~Kouzelis et al.}

\institute{}

\maketitle

\begin{figure*}[th]
  \centering
  \setlength{\tabcolsep}{0.3pt}
  \setlength{\fboxsep}{0pt}
  \renewcommand{\arraystretch}{1.0}

  \newcommand{\phimg}{%
    \fbox{\rule{0pt}{1.1cm}\rule{1.1cm}{0pt}}%
  }
  \newcommand{\phimgteal}{%
    {\color{teal}\setlength{\fboxrule}{1.2pt}\fbox{\rule{0pt}{1.1cm}\rule{1.1cm}{0pt}}}%
  }

  \begin{tabular}{@{}c@{\hspace{3pt}}cccccc@{}}
    \includegraphics[width=1.4cm,height=1.4cm]{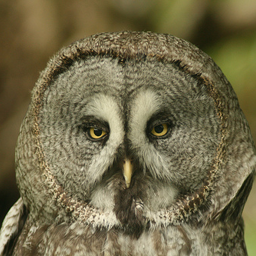} & \includegraphics[width=1.4cm,height=1.4cm]{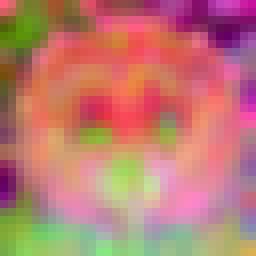} & \includegraphics[width=1.4cm,height=1.4cm]{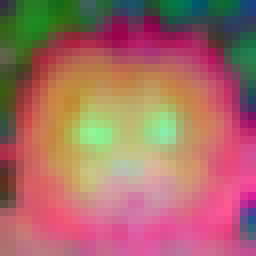} & \includegraphics[width=1.4cm,height=1.4cm]{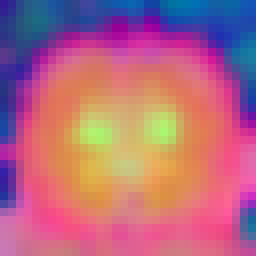} & \includegraphics[width=1.4cm,height=1.4cm]{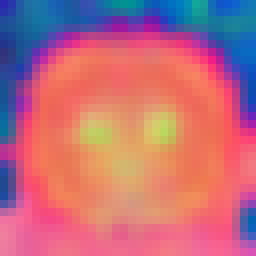} & \includegraphics[width=1.4cm,height=1.4cm]{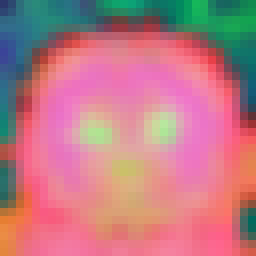} & \includegraphics[width=1.4cm,height=1.4cm]{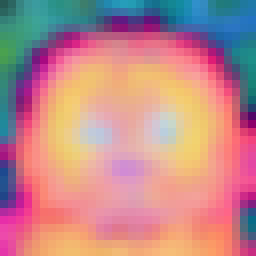} \\
    \includegraphics[width=1.4cm,height=1.4cm]{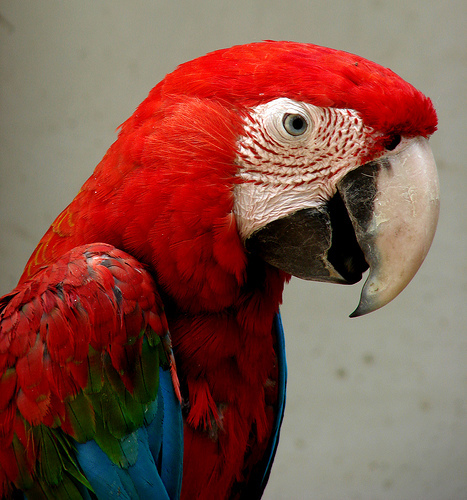} & \includegraphics[width=1.4cm,height=1.4cm]{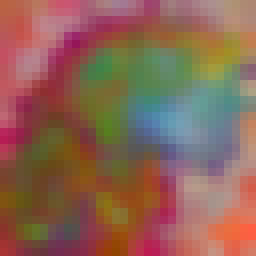} & \includegraphics[width=1.4cm,height=1.4cm]{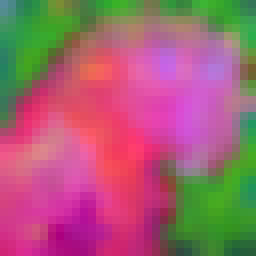} & \includegraphics[width=1.4cm,height=1.4cm]{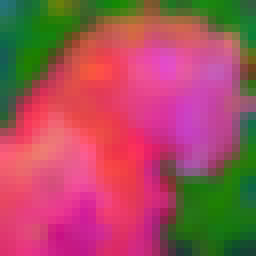} & \includegraphics[width=1.4cm,height=1.4cm]{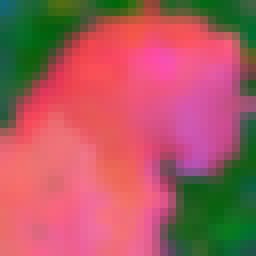} & \includegraphics[width=1.4cm,height=1.4cm]{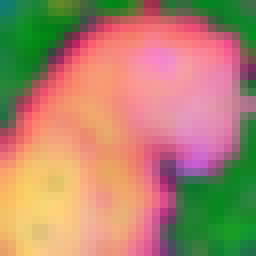} & \includegraphics[width=1.4cm,height=1.4cm]{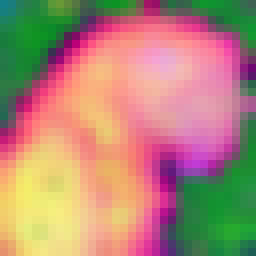} \\
    \includegraphics[width=1.4cm,height=1.4cm]{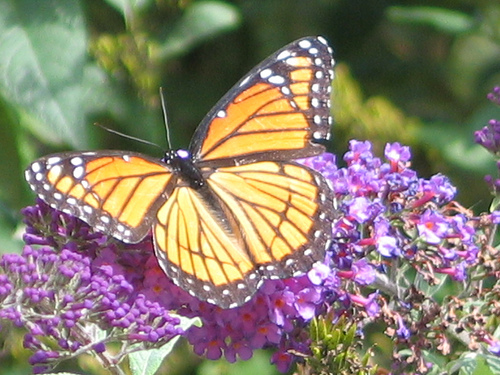} & \includegraphics[width=1.4cm,height=1.4cm]{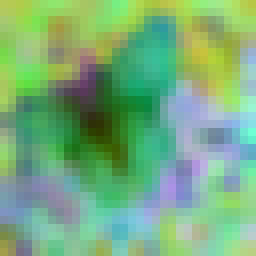} & \includegraphics[width=1.4cm,height=1.4cm]{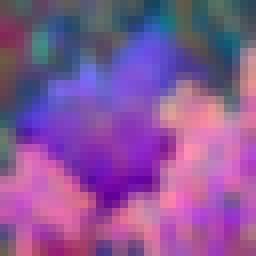} & \includegraphics[width=1.4cm,height=1.4cm]{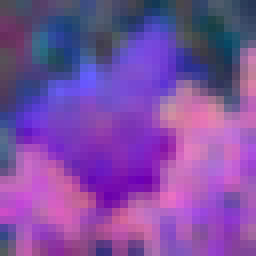} & \includegraphics[width=1.4cm,height=1.4cm]{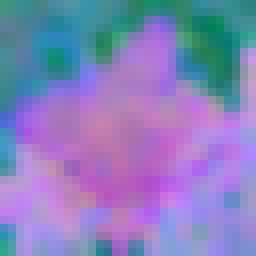} & \includegraphics[width=1.4cm,height=1.4cm]{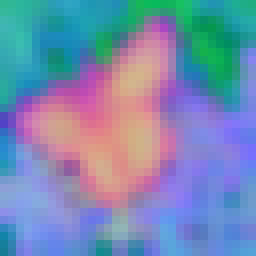} & \includegraphics[width=1.4cm,height=1.4cm]{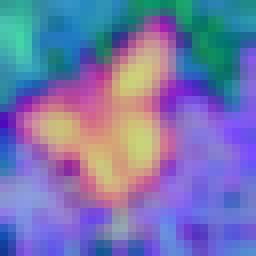} \\
  \end{tabular}

\begin{tikzpicture}

  \hspace{22pt}\draw[->, line width=1.2pt] (2,0) -- (10.5cm, 0);
  \node[anchor=north] at (6.15cm, 0cm) {\small Diffusion Training Steps};
\end{tikzpicture}

  \vspace{2pt}

  \caption{Evolution of the representations throughout \texttt{CoReDi} training. As training progresses, the coevolving representations develop increasingly structured and semantically meaningful spatial organization.}
  \label{fig:teaser}
\end{figure*}
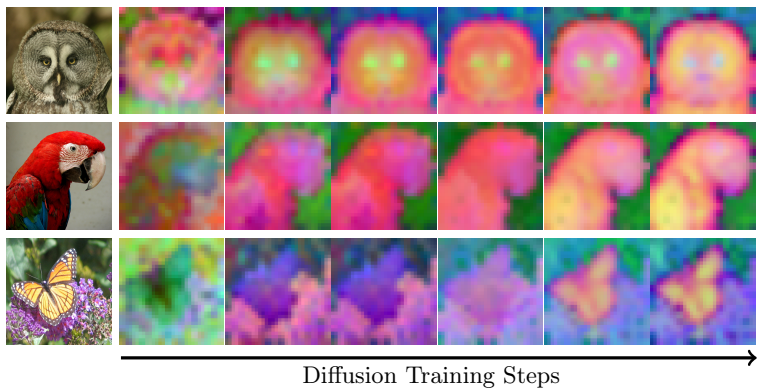\begin{figure*}[t]

\end{figure*}

\vspace{-30pt}
\begin{abstract}

Joint image–feature generative modeling has recently emerged as an effective strategy for improving diffusion training by coupling low-level VAE latents with high-level semantic features extracted from pretrained visual encoders. However, existing approaches rely on a fixed representation space, constructed independently of the generative objective and kept unchanged during training. We argue that the representation space guiding diffusion should itself adapt to the generative task. To this end, we propose Coevolving Representation Diffusion (\texttt{CoReDi}), a framework in which the semantic representation space evolves during training by learning a lightweight linear projection jointly with the diffusion model. While naïvely optimizing this projection leads to degenerate solutions, we show that stable coevolution can be achieved through a combination of stop-gradient targets, normalization, and targeted regularization that prevents feature collapse. This formulation enables the semantic space to progressively specialize to the needs of image synthesis, improving its complementarity with image latents. We apply \texttt{CoReDi} to both VAE latent diffusion and pixel-space diffusion, demonstrating that adaptive semantic representations improve generative modeling across both settings. Experiments show that \texttt{CoReDi} achieves faster convergence and higher sample quality compared to joint diffusion models operating in fixed representation spaces. 

\end{abstract}

\section{Introduction}

Diffusion models~\cite{ddpm, dhariwal2021diffusion, rombach2022high} have become the dominant paradigm for high-fidelity image synthesis. Most modern systems operate either in pixel space or in compressed \texttt{VAE} latent spaces, modeling low-level image statistics with remarkable precision. However, they do not explicitly leverage the rich semantic structure captured by large pretrained visual encoders.

Recent work has explored incorporating semantic priors into diffusion. Approaches include aligning pretrained representations with VAE latents or intermediate diffusion features~\cite{yu2025repa, leng2025repae, singh2025matters, yao2025reconstruction}, replacing VAE latents with such pretrained representations~\cite{zheng2025diffusion, tong2026scaling, shi2025latent}, or \emph{jointly modeling} low-level VAE latents and high-level semantic representations within a unified diffusion process~\cite{kouzelis2025boosting, wu2025representation, petsangourakis2025reglue, semvae}. 

In this latter joint modeling paradigm, the semantic representation space serves as an auxiliary high-level space that complements the low-level \texttt{VAE} latents, forcing the generative model to capture both precise local details (via the \texttt{VAE} latents) and semantic structure (via the representation space). However, in existing joint approaches, the semantic representation space—over which the joint diffusion process operates—is constructed independently of the generative objective and remains fixed during training. In practice, \texttt{PCA} or lightweight autoencoders are used to project the typically high-dimensional semantic features into a more compact space (i.e., with fewer channels) \cite{semvae}. The diffusion model is then trained to learn the distribution of this predefined projection space. Crucially, the representation space itself is not optimized for the generative objective.

This raises a fundamental question: \emph{Should the representation space guiding diffusion remain fixed, or should it adapt jointly with the generative model?}

\subsubsection{Coevolving Representation Diffusion.}

We introduce \emph{Coevolving Representation Diffusion} (\texttt{CoReDi}), a framework in which the projection of pretrained visual features is learned jointly with the diffusion model. Instead of using a predefined mapping (e.g.\ \texttt{PCA}), we train a learnable projection from a frozen visual encoder whose output coevolves with the generative model under the joint diffusion objective. The representation space is therefore no longer an externally imposed target, but a learnable component optimized directly for image synthesis, as illustrated by the evolution of the learned representations throughout training in \autoref{fig:teaser}.

Naïvely optimizing the projection with the joint diffusion loss leads to degenerate solutions, since both the representation input and its denoising target become trainable. Through systematic analysis, we identify three necessary ingredients for stable coevolution:
\begin{enumerate}
    \item \textbf{Stop-gradient in the representation diffusion target}, preventing trivial minimization of the representation diffusion loss.
    \item \textbf{Batch normalization after the projection}, which stabilizes feature scale, preserves the intended noise schedule, and avoids per-channel sample collapse.
\item \textbf{Explicit regularization against feature collapse.} Even with stop-gradient and batch normalization, we observe \emph{feature collapse}, where multiple channels encode redundant information or fail to capture meaningful variation. To address this, we introduce explicit regularization terms that enforce diversity and information preservation in the projected space. We explore simple yet effective strategies, including feature-variance regularization, orthogonality constraints on the projection weights, and covariance regularization to discourage feature collapse. These regularizers play a central role in ensuring that the learned representation remains expressive and complementary to the image latents.

\end{enumerate}
We demonstrate empirically that all three components are necessary for effective coevolution.

\subsubsection{Beyond VAE Latents.}  
Finally, we ask whether joint image–feature diffusion must rely on \texttt{VAE} latents at all. While \texttt{VAE} latents provide computational efficiency, they introduce a reconstruction bottleneck that may limit ultimate image fidelity. Since the auxiliary semantic space already enforces high-level structure, we show that \texttt{CoReDi} extends naturally to pixel-space diffusion, removing the reconstruction bottleneck imposed by \texttt{VAE} compression. 
Building on \texttt{DeCo}~\cite{ma2025deco}, we develop an efficient pixel-space variant in which pretrained visual features coevolve jointly with raw pixels during training, yielding substantial improvements over the baseline \texttt{DeCo} pixel diffusion model and demonstrating that adaptive semantic guidance remains beneficial beyond latent-based generation.

\subsubsection{Contributions.} In summary, our contributions are as follows:
\begin{itemize}
    \item We propose \texttt{CoReDi}, a framework for jointly modeling images and semantic representations in which the representation space itself coevolves with the diffusion model.
    \item We identify and analyze three necessary ingredients for stable training, highlighting the critical role of explicit regularization in preventing feature collapse.
    \item We show that coevolving representations improve joint diffusion in both \texttt{VAE} latent space and pixel space (see \autoref{fig:teaser_page_2}).
\end{itemize}

\definecolor{coral}{HTML}{E07B6A}
\definecolor{softorange}{HTML}{E08C3A}
\definecolor{dustypurple}{HTML}{7B6FA0} 

\definecolor{mmagenta}{HTML}{C94F7C}
\definecolor{eolive}{HTML}{7A9E3B}

\begin{figure*}[tb]
\centering

\begin{minipage}[t]{0.37\textwidth}
  \vspace{0pt}%
  \centering
  \label{fig:teaser_latent}
  \footnotesize

  \newcommand{\myfigA}[1]{\includegraphics[width=0.258\textwidth,valign=c]{#1}}

  \setlength{\tabcolsep}{1pt}

\begin{tabular}{@{}c@{\hspace{2pt}}|@{\hspace{2pt}}cc@{\hspace{2pt}}|@{\hspace{2pt}}cc@{}}
    {\small \texttt{\textbf{Image}}} &
    {\small  \texttt{\textbf{DINOv2}}} &
    {\small \makecell{\texttt{\textbf{+CoReDi}}}} &
    {\small \texttt{\textbf{MOCOv3}}} &
    {\small \makecell{ \texttt{\textbf{+CoReDi}}}} \\

    \myfigA{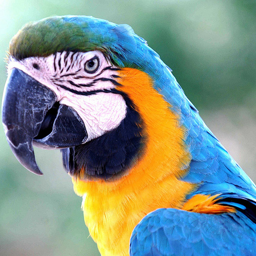} &
    \myfigA{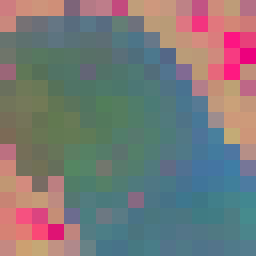} &
    \myfigA{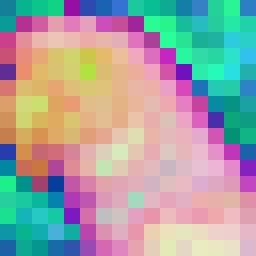} &
    \myfigA{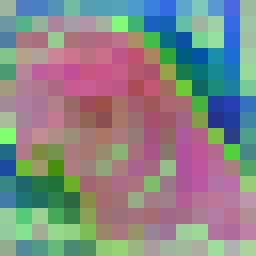} &
    \myfigA{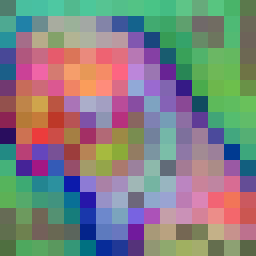} \\

    \multicolumn{5}{c}{\vspace{-2.1ex}}\\

    \myfigA{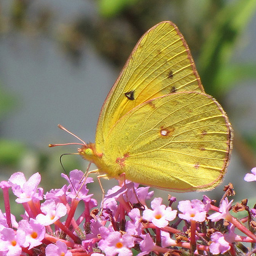} &
    \myfigA{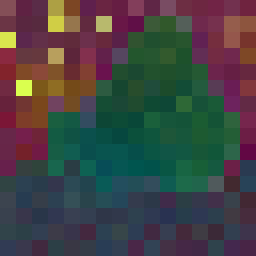} &
    \myfigA{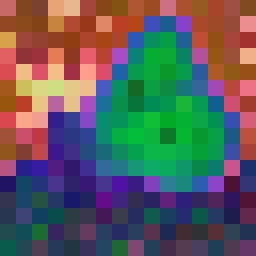} &
    \myfigA{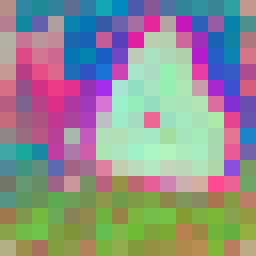} &
    \myfigA{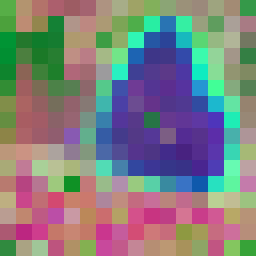} \\

    \multicolumn{5}{c}{\vspace{-2.1ex}}\\

    \myfigA{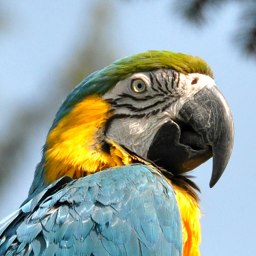} &
    \myfigA{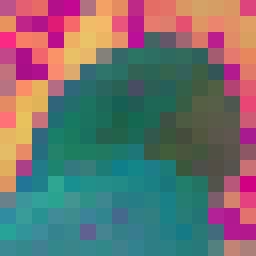} &
    \myfigA{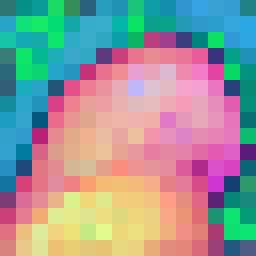} &
    \myfigA{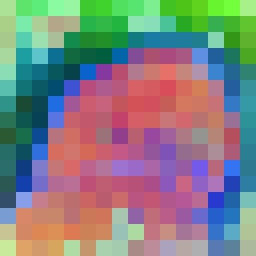} &
    \myfigA{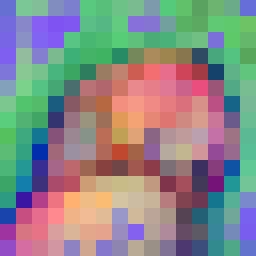} \\

    \multicolumn{5}{c}{\vspace{-2.1ex}}\\

    \myfigA{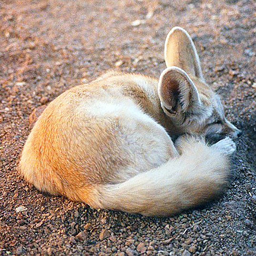} &
    \myfigA{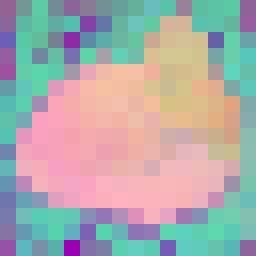} &
    \myfigA{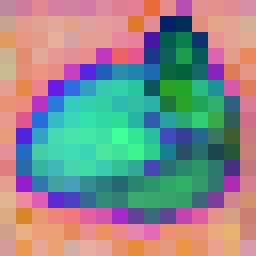} &
    \myfigA{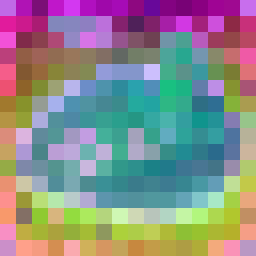} &
    \myfigA{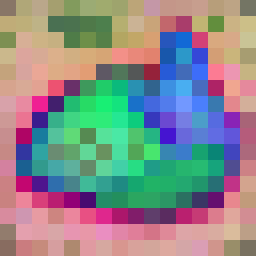} \\

  \end{tabular}


\vspace{6pt}
\makebox[\textwidth][c]{
  \hspace*{6.4cm}\parbox{0.7\textwidth}{
  }
}

\end{minipage}
\hfill
\begin{minipage}[t]{0.49\textwidth}
  \vspace{0pt}%
  \centering
  \footnotesize
\begin{tikzpicture}
    \begin{axis}[
      scale only axis,
      width=0.7\linewidth,
      height=0.35\linewidth,
      xmin=85000, xmax=510000,
      ymin=4.5, ymax=25,
      xtick={100000, 200000, 310000, 420000, 500000},
      xticklabels={100K, 200K, 300K, 400K, 4M},
      ytick={5,10,15,20,25,30},
      yticklabels={5,10,15,20,25,40},
      scaled x ticks=false,
      ylabel={FID},
      tick label style={font=\scriptsize},
      label style={font=\scriptsize},
      legend style={
          draw=black,
          line width=0.5pt,
          at={(0.99,0.99)},
          font=\scriptsize,
          anchor=north east
      },
      grid=major,
      major grid style={gray!20},
    ]
      \addplot[color=teal!20, mark=*, line width=1.2pt]
        coordinates {
           (500000, 5.9)
        };
      \addlegendentry{\texttt{REPA-XL/2}}
      \addplot[color=teal!40, mark=*, line width=1.2pt]
        coordinates {
           (100000, 23.1)
           (200000, 12.6)
           (310000, 9.5)
           (420000, 7.5)
        };
      \addlegendentry{\texttt{ReDi-XL/2}}
      \addplot[color=teal, mark=*, line width=1.2pt]
        coordinates {
           (100000, 17.1)
           (200000, 9.4)
           (310000, 6.10)
        };
      \addlegendentry{\texttt{CoReDi-XL/2}}

      \draw[<-, dashed, teal]
        (axis cs:320000, 6) -- (axis cs:500000, 6)
        node[pos=0.75, below=-4mm] {\(\times 13\) };

    \end{axis}
\end{tikzpicture}

\vspace{4pt}

\begin{tikzpicture}
    \begin{axis}[
      scale only axis,
      width=0.7\linewidth,
      height=0.35\linewidth,
      xmin=40, xmax=210,
      ymin=16, ymax=92,
      xtick={50, 100, 150, 200},
      xticklabels={50K, 100K, 150K, 200K},
      ytick={18,32,50,70,90},
      yticklabels={20,35,50,70,90},
      scaled x ticks=false,
      ylabel={FID},
      tick label style={font=\scriptsize},
      label style={font=\scriptsize},
      legend style={
          draw=black,
          line width=0.5pt,
          at={(0.99,0.99)},
          font=\scriptsize,
          anchor=north east
      },
      grid=major,
      major grid style={gray!20},
    ]

      \addlegendentry{\texttt{DeCo-L/16}}
      \addplot[color=dustypurple!40, mark=*, line width=1.2pt]
        coordinates {
           (50, 88)
           (100, 48)
           (150, 37)
           (200, 31)
        };
      \addlegendentry{\texttt{CoReDi-L/16}}
      \addplot[color=dustypurple, mark=*, line width=1.2pt]
        coordinates {
           (50, 72)
           (100, 31.1)
           (150, 22)
           (200, 19)
        };

      \draw[<-, dashed, dustypurple]
        (axis cs:110, 31) -- (axis cs:200, 31)
        node[pos=0.01, below=0.12mm] {\(\times 2\) };

    \end{axis}
\end{tikzpicture}

\vspace{0.5pt}
\makebox[\textwidth][c]{
  \hspace*{6.4cm}\parbox{0.7\textwidth}{
  }
}
\end{minipage}
\vspace{-5pt}

\caption{(Left) Comparison of fixed PCA and learned \texttt{CoReDi} representations for \texttt{DINOv2} and \texttt{MOCOv3}. The learned projections yield cleaner, more structured representations with coherent spatial organization, while the fixed PCA projections produce noisier, less semantically meaningful activations. (Right) By jointly adapting the representation space alongside the generative model, \texttt{CoReDi} consistently speeds up convergence. In latent space (Top), \texttt{CoReDi} outperforms \texttt{ReDi} and, notably, converges $\sim 13\times$ faster than \texttt{REPA}. In pixel space (Bottom) it improves convergence by $\times 2$ over \texttt{DeCo}.}
\label{fig:teaser_page_2}
\end{figure*}
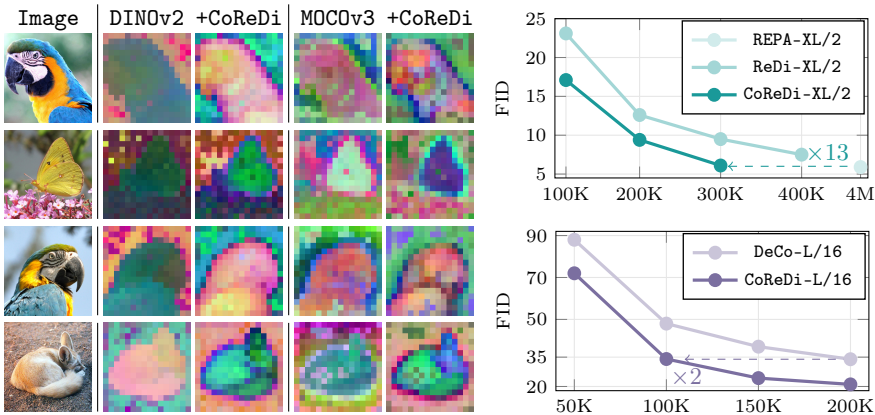

\section{Related Work}

\begingroup\sloppy\emergencystretch=4em

\subsubsection{Latent Diffusion Models.} 
Latent Diffusion Models~\cite{rombach2022high, ma2024sit, peebles2023scalable, zheng2024masked, wang2025ddt} operate in the compressed latent space of a variational autoencoder (\texttt{VAE})~\cite{rombach2022high, yao2025reconstruction, kouzelis2025eqvae}, which reduces spatial dimensionality compared to pixel-space diffusion, significantly lowering computational cost and learning difficulty~\cite{rombach2022high}. The Diffusion Transformer (\texttt{DiT})~\cite{peebles2023scalable} marked a significant architectural shift by replacing the U-Net \cite{ronneberger2015u} backbone with a transformer, and SiT~\cite{ma2024sit} extended this framework to flow-based diffusion objectives. 

\subsubsection{Pixel Space Diffusion.} 
Recently, there has been a surge of interest in pixel diffusion models, since they avoid the reconstruction bottleneck imposed by the \texttt{VAE}. Early approaches relied on multi-stage pipelines operating at progressively increasing resolutions to manage the high dimensionality of pixel space~\cite{teng2023relay, chen2025pixelflow}, at the expense of more complex training and inference procedures. More recent work has explored alternative architectures to sidestep these issues, including transformer-based normalizing flows~\cite{zhai2024normalizing}, fractal generative models~\cite{li2025fractal}, \texttt{DiT}-based models that predict neural field parameters per patch~\cite{wang2025pixnerd}, and methods that predict the clean image directly to anchor generation to the low-dimensional data manifold~\cite{li2025back}. \texttt{DeCo}~\cite{ma2025deco} decouples the generation of high and low frequency components, leveraging a lightweight pixel decoder to reduce the complexity of direct pixel synthesis. Despite these advances, the integration of visual representations into pixel diffusion models to enhance generative performance remains largely unexplored.

\subsubsection{Semantic Representations in Generative Models.}
Recent work has explored leveraging semantic representations\cite{oquab2023dinov2,ma2025deco,Chen2021AnES,tschannen2025siglip,venkataramanan2025franca} to enhance generative modeling  ~\cite{yu2025repa,zheng2025diffusion, shi2025latent, chen2025aligning, Shi2025LatentDM, tong2026scaling, kouzelis2025boosting, wu2025representation, karypidis2024dino, petsangourakis2025reglue, semvae}. \texttt{REPA}~\cite{yu2025repa} aligns diffusion features with pretrained visual encoders, while \texttt{REPA-E}~\cite{leng2025repae} enables end-to-end joint optimization of the \texttt{VAE} and diffusion model, and \texttt{iREPA}~\cite{singh2025matters} improves the spatial structure of the representation used for alignment. Many recent works ~\cite{zheng2025diffusion, shi2025latent, chen2025aligning, Shi2025LatentDM, tong2026scaling} replace the \texttt{VAE} with pretrained visual encoder representations, improving generation. However, by keeping the encoder frozen, these methods fall behind state-of-the-art \texttt{VAE}s in reconstruction quality~\cite{bfl2025representation}. \texttt{REG}~\cite{wu2025representation} and \texttt{ReDi}~\cite{kouzelis2025boosting} jointly model low-level \texttt{VAE} features and high-level semantic features from \texttt{DINOv2}~\cite{oquab2023dinov2}, with \texttt{ReDi} using \texttt{PCA}-compressed patch embeddings. Instead of relying on static representations, we propose allowing the input representation to evolve dynamically during training, further improving generation quality.

\subsubsection{Preventing Representation Collapse in Self-Supervised Learning.}
Self-supervised approaches are often prone to representational collapse, and several mechanisms have been proposed to address this. Redundancy-reduction methods such as \texttt{Barlow Twins}~\cite{zbontar2021barlow}, \texttt{VICReg}~\cite{bardes2021vicreg}, and \texttt{W-MSE}~\cite{ermolov2021whitening} decorrelate features to avoid degenerate solutions, with \texttt{VICRegL}~\cite{bardes2022vicregl} extending this to local features. \texttt{SimCLR}~\cite{chen2020simple} highlights batch normalization as an implicit collapse prevention mechanism, coupling examples within a batch to discourage trivial constant representations. Architectural approaches, \texttt{BYOL}~\cite{grill2020bootstrap}, \texttt{SimSiam}~\cite{chen2021exploring}, and \texttt{DINO}~\cite{Caron2021EmergingPI, oquab2023dinov2} instead break gradient symmetry via stop-gradients, momentum encoders, or output centering.

\endgroup

\label{sec:related}

\section{Method}
\label{sec:method}

In this section, we describe the \texttt{CoReDi} framework for jointly modeling images and coevolving semantic representations under a diffusion objective. We begin by reviewing the preliminary setup of joint image-feature synthesis (\cref{sec:preliminary}). Next, we introduce our learnable projection, allowing the semantic representation space to evolve alongside the generative model, including stabilization techniques such as batch normalization and stop-gradient (\cref{sec:coevolving}). We then discuss explicit regularization strategies designed to prevent feature collapse and ensure diversity in the learned representations (\cref{sec:regulizers}). Finally, we present the overall training objective (\cref{sec:overall_training}) and describe a natural extension of \texttt{CoReDi} to pixel-space diffusion (\cref{sec:coevolve_pixel}).

\subsection{Preliminary: Joint Image-Feature Synthesis} \label{sec:preliminary}

\subsubsection{Joint Flow Matching Objective.}
The joint image–feature generation framework of \cite{kouzelis2025boosting} trains a single flow matching model~\cite{albergo2025stochastic, esser2024scaling, lipman2022flow}  to jointly capture low-level image structure and high-level semantic information. Given an image latent $\mathbf{x}_0 \sim p(\mathbf{x})$ and its corresponding visual representation $\mathbf{z}_0=\texttt{VE}(\mathbf{x}_0) \in \mathbb{R}^{L\times D}$ extracted by a frozen pretrained encoder $\texttt{VE}$ where $L$ is the number of spatial tokens and $D$ is the feature dimension. Since the dimensionality of the semantic features $D$ greatly exceeds that of the image latents, \cite{kouzelis2025boosting} reduces it via a fixed PCA projection $\mathbf{P} \in \mathbb{R}^{D\times d}$, computed once prior to training, yielding $\tilde{\mathbf{z}}_0 = \mathbf{z}_0 \mathbf{P} \in \mathbb{R}^{L\times d}$. Using $\mathbf{x}_0,\tilde{\mathbf{z}}_0$ with $d\ll D$, the coupled interpolation process is defined as

\begin{equation}
\label{eq:joint_forward_proc_redi}  
    \mathbf{x}_t = (1-t) \mathbf{x}_0 + t \boldsymbol{\epsilon}_x, \quad  
    \tilde{\mathbf{z}}_t = (1-t) \tilde{\mathbf{z}}_0 + t \boldsymbol{\epsilon}_z, 
\end{equation}  
A network $\mathbf{v}_\theta(\mathbf{x}_t,\tilde{\mathbf{z}}_t,t)$ then predicts the velocities for both modalities via two heads, $\mathbf{v}^x_\theta(\mathbf{x}_t,\tilde{\mathbf{z}}_t,t)$ and $\mathbf{v}^z_\theta(\mathbf{x}_t,\tilde{\mathbf{z}}_t,t)$. Training minimizes the joint flow-matching objective:
\begin{equation}  
\label{eq:redi_loss}
    \mathcal{L}_{\text{joint}}(\mathbf{x}_0, \tilde{\mathbf{z}}_0, t) = \underbrace{\Vert \mathbf{v}^x_\theta(\mathbf{x}_t,\tilde{\mathbf{z}}_t, t) - 
    (\boldsymbol{\epsilon}_x -
    \mathbf{x}_0)
    \Vert^2}_{\mathcal{L}_{\text{image}}}  
    + \lambda_z \underbrace{\Vert \mathbf{v}^z_\theta(\mathbf{x}_t,\tilde{\mathbf{z}}_t, t) - 
    (\boldsymbol{\epsilon}_z -
    \tilde{\mathbf{z}}_0)
    \Vert^2}_{\mathcal{L}_{\text{rep}}},
\end{equation}  
where $\lambda_z$ balances the contribution of the representation loss $\mathcal{L}_{\text{rep}}$ relative to the image loss $\mathcal{L}_{\text{image}}$.

\subsubsection{Merged Tokens Strategy.}
We adopt the merged tokens strategy \cite{kouzelis2025boosting} to fuse image and representation. Both modalities are embedded separately and summed channel-wise, $\mathbf{h}_t = \mathbf{x}_t \mathbf{W}_{\text{emb}}^x + \tilde{\mathbf{z}}_t \mathbf{W}_{\text{emb}}^z$, before being processed by the transformer. Separate predictions for each modality are then obtained via modality-specific decoding heads, $\mathbf{v}^x_\theta = \mathbf{o}_t \mathbf{W}_{\text{dec}}^x$ and $\mathbf{v}^z_\theta = \mathbf{o}_t \mathbf{W}_{\text{dec}}^z$ where $\mathbf{o}_t$ is the output of the diffusion transformer. This early fusion approach enables joint modeling of both modalities while preserving the original token count, incurring no additional computational overhead over the standard diffusion transformer.

\begin{figure}[t]
        \centering
        \includegraphics[width=0.9\textwidth]{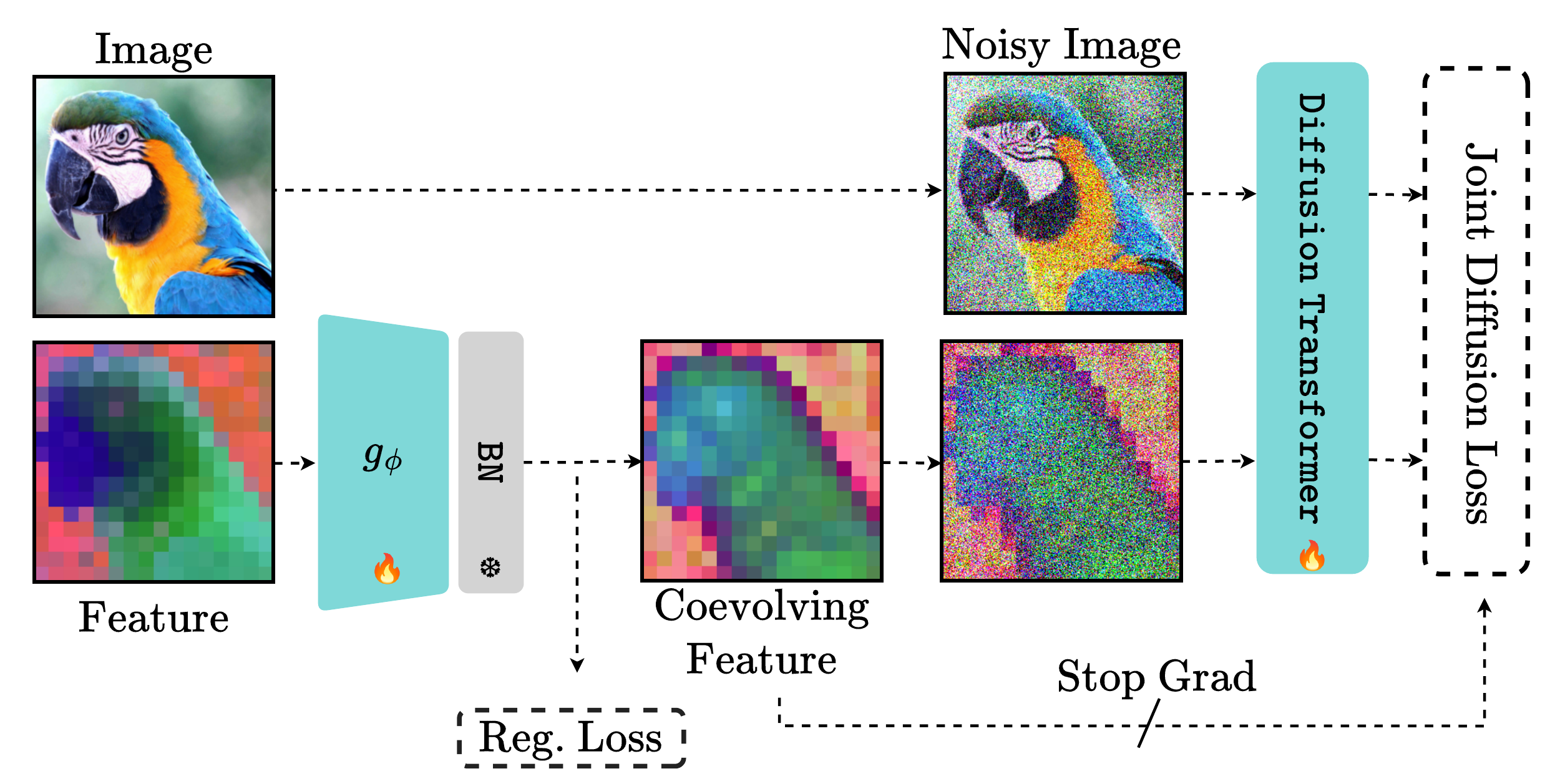} 

    \caption{\textbf{Overview of CoReDi}. Given an input image, a frozen pretrained visual encoder extracts semantic features, which are projected to a lower-dimensional space via a learnable projection $g_\phi$, followed by batch normalization and a regularization loss to prevent collapse. Both the noisy image tokens and the noisy coevolving feature tokens are passed as input to a diffusion backbone, which jointly predicts the image and representation velocities. A stop-gradient is applied through the clean representation target in the representation loss, allowing the projection to coevolve with the generative model without degeneracy.}
    \label{fig:method}
    \vspace{-10pt}
\end{figure}

\subsection{Coevolving Representation Diffusion} \label{sec:coevolving}

While \cite{kouzelis2025boosting} reduces the dimensionality of the pretrained visual representation using a \emph{fixed} PCA projection, we instead learn an adaptive projection $g_\phi(\cdot)$ \emph{jointly} with the generative model. Concretely, given the frozen encoder output $\mathbf{z}_0=\texttt{VE}(\mathbf{x}_0)$, we replace the fixed mapping with a learnable projection:
\begin{equation}
\label{eq:learnable_proj}
\tilde{\mathbf{z}}_0 = g_\phi (  \mathbf{z}_0 ),
\end{equation}
Unlike the fixed PCA projection, $g_\phi$ adapts throughout training, allowing the representation space to evolve alongside the generative model to better assist image synthesis. In this work, we explore $g_\phi$ as a simple trainable linear layer  $g_\phi(\mathbf{z}_0) = \mathbf{z}_0\mathbf{W_\phi}$ with $\mathbf{W_\phi} \in \mathbb{R}^{D\times d}$ where D is the feature dimension of \texttt{VE}.

\subsubsection{Batch Normalization.} Diffusion models are highly sensitive to input scale, as variations in feature statistics implicitly distort the intended noise schedule and destabilize training. To mitigate this, we apply batch normalization after the learnable projection using exponential moving average estimates of the mean and variance. Beyond scale stabilization, batch normalization acts as an implicit regularizer against sample collapse, enforcing a non-degenerate distribution over samples at each feature channel. We omit the standard trainable affine parameters (scale and shift), as the purpose of normalization here is solely to control input scale and prevent collapse, rather than to allow the network to rescale or shift the normalized features.

\definecolor{dustydustypurple}{HTML}{7B6FA0} 
\subsubsection{Stop-Gradient.}
Directly optimizing the projection $g_\phi$  via \autoref{eq:redi_loss} leads to degenerate solutions since both the input to the model and the target are trainable. To stabilize training, we stop gradients through the \emph{clean} projected target used in the representation velocity loss:
\begin{equation} \label{eq:eredi_loss}  
   \mathcal{L}_{\text{rep}}(\mathbf{x}_0, \tilde{\mathbf{z}}_0, t) =   
\Vert \textcolor{black}{\mathbf{v}^z_\theta}(\mathbf{x}_t,\tilde{\mathbf{z}}_t, t) - 
    (\boldsymbol{\epsilon}_z -
    \textcolor{purple}{\texttt{sg}(}\tilde{\mathbf{z}}_0\textcolor{purple}{)})
    \Vert^2 ,  
\end{equation}  
where $\textcolor{purple}{\texttt{sg}(\cdot)}$ denotes the stop-gradient operator. In this way, the diffusion model learns to jointly denoise image and representation tokens, while the representation space can coevolve without allowing the representation target itself to be trivially modified to reduce the loss.

\subsection{Regularization Methods} \label{sec:regulizers}

\begin{figure*}[t]
  \centering
  \setlength{\tabcolsep}{0.2pt}
  \newcommand{\myfigA}[1]{\includegraphics[width=0.1\textwidth]{#1}}
  \newcommand{\myfigB}[1]{\setlength{\fboxrule}{1pt}\setlength{\fboxsep}{0pt}\fbox{\includegraphics[width=0.1\textwidth]{#1}}}
  \newcommand{\vfimg}{\fbox{\rule{0pt}{1.2cm}\rule{1.2cm}{0pt}}}
  \hspace{-20pt}
\begin{tabular}{@{}c@{\hspace{3pt}}cccc@{\hspace{6pt}}cccc@{}}
    \multirow{2}{*}[9pt]{\myfigB{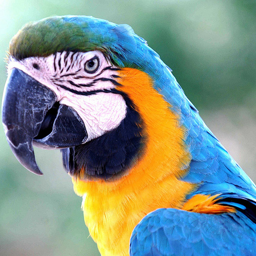}} & \myfigA{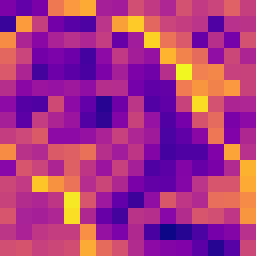} & \myfigA{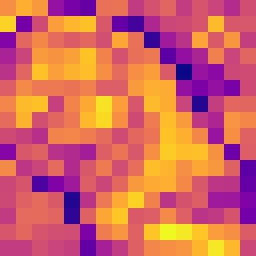} & \myfigA{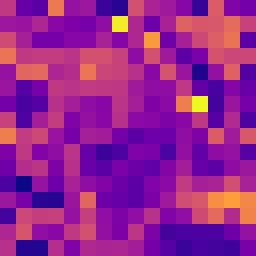} & \myfigA{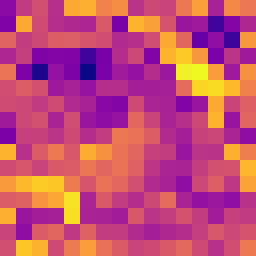} & \myfigA{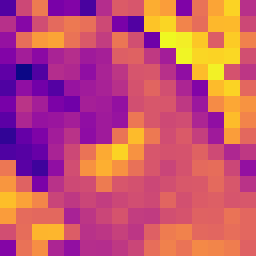} & \myfigA{figs/0_grid_vf1/channel_00.png} & \myfigA{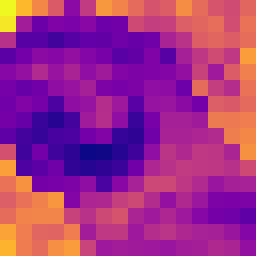} & \myfigA{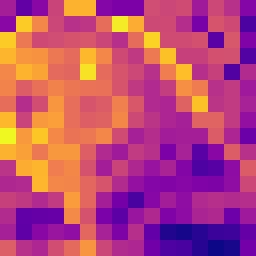} \\ [-2pt]
    & \myfigA{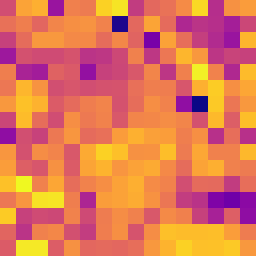} & \myfigA{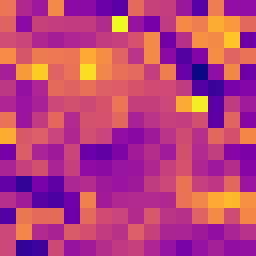} & \myfigA{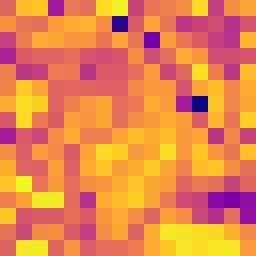} & \myfigA{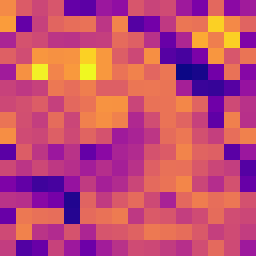} & \myfigA{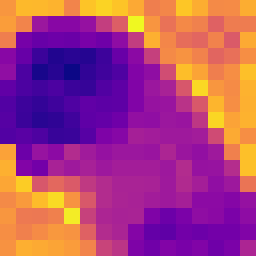} & \myfigA{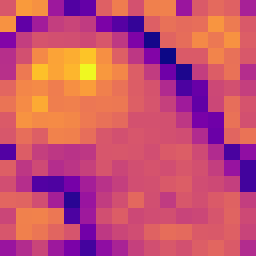} & \myfigA{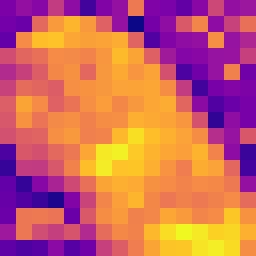} & \myfigA{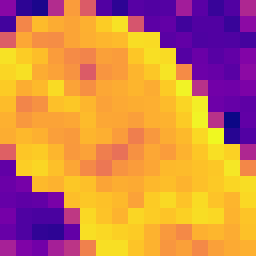} \\
    
    \multirow{2}{*}[9pt]{\myfigB{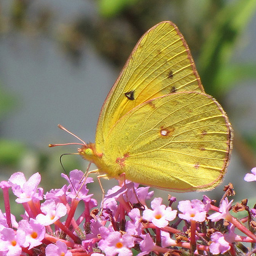}} & \myfigA{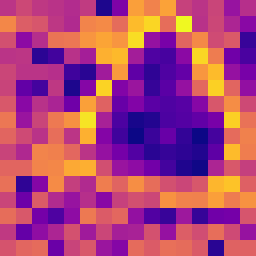} & \myfigA{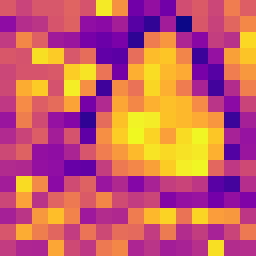} & \myfigA{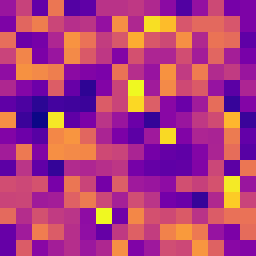} & \myfigA{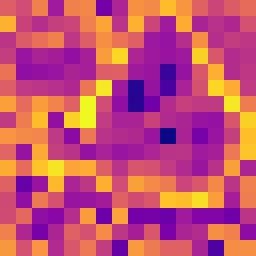} & \myfigA{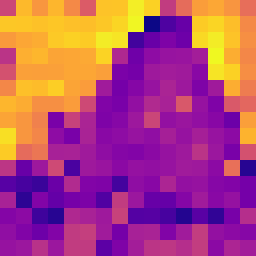} & \myfigA{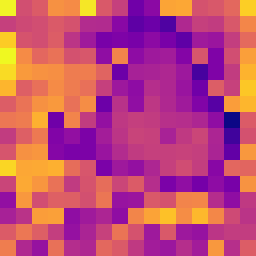} & \myfigA{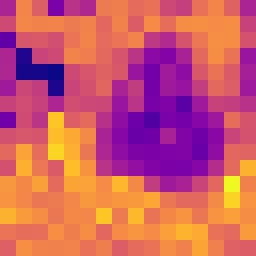} & \myfigA{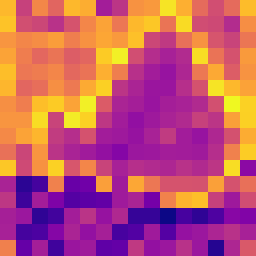} \\ [-2pt]
    & \myfigA{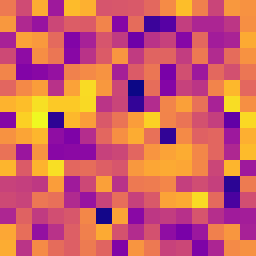} & \myfigA{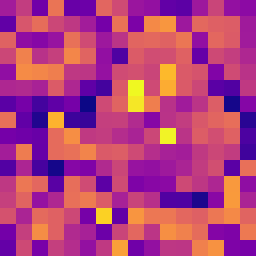} & \myfigA{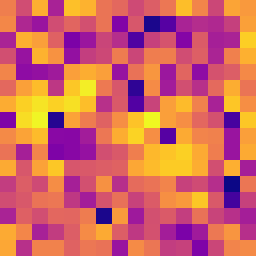} & \myfigA{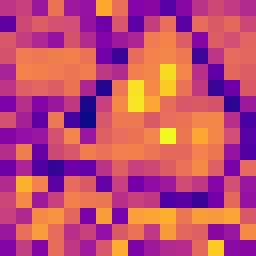} & \myfigA{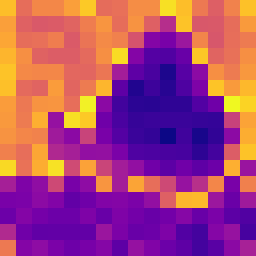} & \myfigA{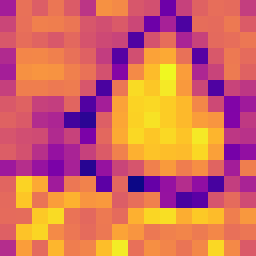} & \myfigA{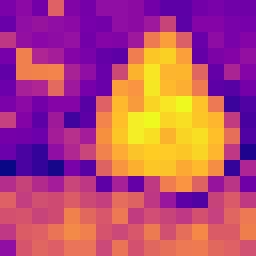} & \myfigA{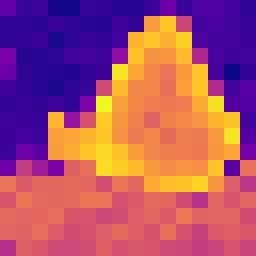} \\ [-3pt]
    
    & \multicolumn{4}{@{}c@{\hspace{6pt}}}{\footnotesize w/o Regularization} & \multicolumn{4}{c@{}}{\footnotesize Feature  Variance Regularization} \\
  \end{tabular}
  \caption{\textbf{Regularization Prevents Feature Collapse.} Visualization of all $8$ channels of the coevolving representation $\tilde{\textbf{z}_0}$ at $200$K steps, under two training configurations. Without regularization, the projected channels collapse, failing to capture diverse semantic information. The Feature Variance Regularization strategy successfully prevents collapse, yielding semantically meaningful channel activations.}
  \label{fig:vf_grid}
\end{figure*}


Batch normalization implicitly prevents \emph{sample collapse}, where different images or spatial locations are projected to the same feature point. However, we observe that learned projections can still exhibit \emph{feature collapse}, where individual feature channels fail to vary meaningfully across samples or fail to carry variation that is distinct from other channels (\autoref{fig:vf_grid}, top left). To address this, we explore the following regularization strategies.



\subsubsection{Feature Variance Regularization.}  
To prevent feature collapse, we encourage each channel of the representation to exhibit sufficient variation. Unlike \texttt{VICReg}~\cite{bardes2021vicreg}, which enforces variance across the batch dimension, we instead consider each feature vector $\tilde{\mathbf{z}}_0^i$ (at location $i$) and penalize feature vectors whose standard deviation across the channel dimension falls below a threshold $\gamma$ via a hinge loss:
\[
\mathcal{L}_{\text{var}}(\tilde{\mathbf{z}}_0) = \frac{1}{L} \sum_{i=1}^{L} \max\left(0, \gamma - \sqrt{\text{Var}(\tilde{\mathbf{z}}_0^i) + \epsilon}\right),
\]
where $L$ is the number of spatial tokens, $\epsilon$ ensures numerical stability, and $\gamma = 1$ sets the minimum desired standard deviation. This encourages each channel to remain active and carry meaningful variation, preventing feature collapse and redundancy across channels.

\subsubsection{Orthogonality Regularization.} As an alternative to penalizing feature collapse at the feature level, we instead regularize the weight matrix $\mathbf{W}_\phi$ of the linear projection $\tilde{\mathbf{z}}_0 = \mathbf{z}_0 \mathbf{W}_{\phi}$ directly. Concretely, we penalize the deviation of $\mathbf{W}_\phi^\top\mathbf{W}_\phi$ from the identity matrix:
\[
\mathcal{L}_{\text{orth}} = \left\| \mathbf{W}^\top_\phi\mathbf{W}_\phi - \mathbf{I} \right\|_F^2,
\]
where $\|\cdot\|_F$ denotes the Frobenius norm. By enforcing orthonormality of the projection columns, this regularization structurally prevents feature redundancy, encouraging each projection direction to capture a distinct component of the representation space.

\subsubsection{Covariance Regularization.} Inspired by \cite{zbontar2021barlow, bardes2021vicreg} we penalize the off-diagonal entries of the channel covariance matrix of the projected representations $\tilde{\mathbf{z}}_0$. Concretely, letting $C(\tilde{\mathbf{z}}_0) \in \mathbb{R}^{d \times d}$ denote the normalized channel covariance matrix, we define:
\[
\mathcal{L}_{\text{cov}}(\tilde{\mathbf{z}}_0) = \frac{1}{d} \sum_{i \neq j} \left[ C(\tilde{\mathbf{z}}_0) \right]^2_{i,j},
\]
where $d$ is the number of feature channels. This encourages the off-diagonal entries of $C(\tilde{\mathbf{z}}_0)$ to be close to zero, decorrelating the projected channels and preventing them from encoding redundant information.

\subsection{Overall Training of CoReDi} \label{sec:overall_training}

The training is performed end-to-end by jointly optimizing the diffusion model parameters $\theta$ and the projection parameters $\phi$ as visualized in \autoref{fig:method}. The total training objective is:
\begin{equation}
    \mathcal{L}(\theta, \phi) = \mathcal{L}_{\text{image}}(\theta, \phi) + \lambda_z \mathcal{L}_{\text{rep}}(\theta, \phi) + \lambda_{\text{reg}} \mathcal{L}_{\text{reg}}(\phi),
\end{equation}
where $\mathcal{L}_{\text{image}}$ is the image flow-matching loss, $\mathcal{L}_{\text{rep}}$ is the representation flow-matching loss, and $\mathcal{L}_{\text{reg}}$ is a regularization term applied solely to the projection parameters $\phi$. Specifically, $\mathcal{L}_{\text{reg}}$ can be any of the feature variance, orthogonality, or covariance regularizers, as described in \autoref{sec:regulizers}: $\mathcal{L}_{\text{reg}} \in \{\mathcal{L}_{\text{var}}, \mathcal{L}_{\text{orth}}, \mathcal{L}_{\text{cov}}\}$.  
The hyperparameters $\lambda_z$ and $\lambda_{\text{reg}}$ control the relative contributions of the representation and regularization losses, respectively.

\subsection{Coevolving Representations in Pixel Space} \label{sec:coevolve_pixel}

To extend \texttt{CoReDi} to pixel space, we build upon the encoder-decoder architecture of \texttt{DeCo}~\cite{ma2025deco}, in which a \texttt{DiT} encoder operates on downsampled images and a lightweight pixel decoder reconstructs full-resolution outputs. Given a noisy downsampled image $\hat{\mathbf{x}}_t$ and the noisy coevolving representation $\tilde{\mathbf{z}}_t$, the encoder processes both modalities jointly to produce the joint condition features $\textbf{c}_{\text{joint}}$:
\begin{equation}
    \textbf{c}_{\text{joint}} = \text{Enc}_\theta(\hat{\mathbf{x}}_t, \tilde{\mathbf{z}}_t, t),
\end{equation}
The full-resolution image velocity and representation velocity are then predicted by the pixel decoder and a lightweight linear projection head, respectively:
\begin{equation}
    \mathbf{v}^x_{\text{pred}} = \text{Dec}_\theta(\mathbf{x}_t, \textbf{c}_{\text{joint}}, t), \qquad \mathbf{v}^z_{\text{pred}} = \mathbf{W}_{\text{dec}}\,  \textbf{c}_{\text{joint}}.
\end{equation}
This extension requires only minimal modifications to the \texttt{DeCo} architecture, enabling joint modeling of image and semantic representation tokens within a unified encoder-decoder framework.
\vspace{10pt}

\hspace*{-15pt}
\begin{minipage}[t]{0.38\textwidth}
    \centering
    \fontsize{7.5}{9}\selectfont
    \setlength{\tabcolsep}{2pt}

\begin{tabular}{lccc}
\toprule
\Th{Model} & \Th{\#Params} & \Th{Iter.} & \Th{FID$\downarrow$} \\ 
\midrule 
\texttt{SiT-B/2}   &  $130$M & $400$K & $33.0$   \\
\texttt{ReDi-B/2} & $130$M & $400$K & $21.4$    \\
\rowcolor{teal!10} \texttt{CoReDi-B/2}   & $130$M & $200$K & $24.7$   \\
\rowcolor{teal!10} \texttt{CoReDi-B/2}   & $130$M & $400$K & $16.4$   \\
 
\arrayrulecolor{black!30}\cmidrule(lr){1-4}
\texttt{SiT-XL/2} & $675\text{M}$ & $7\text{M}$ & $8.3$ \\
\texttt{REPA-XL/2}  & $675\text{M}$ & $4\text{M}$ & $5.9$ \\
\texttt{ReDi-XL/2} & $675\text{M}$ & $4\text{M}$ & $3.3$ \\
\rowcolor{teal!10} \texttt{CoReDi-XL/2} & $675\text{M}$ & $2\text{M}$ & $3.3$ \\
\arrayrulecolor{black}\bottomrule
\end{tabular}

\captionsetup{type=table, margin={2pt,2pt}}
\captionof{table}{\textbf{Latent Diffusion Comparison without CFG.} \Th{FID} scores on ImageNet$256$ without Classifier-Free Guidance for \texttt{SiT} models of various sizes with \texttt{REPA}, \texttt{ReDi} and \texttt{CoReDi}.}
\label{tab:fid_comparison}
\end{minipage}
\hspace{3pt}
\begin{minipage}[t]{0.52\textwidth}
\centering

\fontsize{7.5}{9}\selectfont
\setlength{\tabcolsep}{2pt}
\begin{tabular}{l c c c c c c}
\toprule
\Th{Model} & \Th{Epochs} & \Th{FID$\downarrow$} & \Th{sFID$\downarrow$} & \Th{IS$\uparrow$} & \Th{Pre.$\uparrow$} & \Th{Rec.$\uparrow$} \\
\arrayrulecolor{black}\midrule

\rowcolor{TableColorGrey}\multicolumn{7}{c}{\emph{Latent Diffusion Models}} \\
 \texttt{LDM}~\cite{rombach2022high} & 200 & 3.60 & -  & 247.7 & 0.87 & 0.48 \\ 
 \texttt{DiT-XL/2}~\cite{peebles2023scalable}   & 1400  &    2.27 & 4.60 & {278.2} & {0.83} & 0.57  \\
 \texttt{SiT-XL/2}~\cite{ma2024sit}   & 1400 &     2.06 & {4.50} & 270.3 & 0.82 & 0.59 \\
 \arrayrulecolor{black!40}\midrule
 \rowcolor{TableColorGrey}\multicolumn{7}{c}{\emph{Leveraging Visual Representations}} \\
\texttt{REPA}~\cite{yu2025repa} & {800} & {1.80} & {{4.50}} & {{284.0}} & {0.81} & {0.61} \\
\texttt{ReDi}~\cite{kouzelis2025boosting} & 800 & 1.72  & 4.68 & 278.7 & 0.77 & 0.63 \\
\rowcolor{teal!10}\texttt{CoReDi} & \textbf{400} & \textbf{1.58} & {4.33} & {297.2} & {0.63} & 0.78 \\
\arrayrulecolor{black}\bottomrule
\end{tabular}
\captionsetup{type=table, margin={-13pt,2pt}}
\captionof{table}{\textbf{Latent Diffusion Comparison with CFG}. Quantitative evaluation on ImageNet$256$ with Classifier-Free Guidance. \texttt{REPA}, \texttt{ReDi}, and \texttt{CoReDi} all use \texttt{SiT-XL/2} as the base model.}
\label{tab:sota_comparison}
\end{minipage}

\section{Experiments}
\label{sec:exp}

\vspace{-5pt}
\subsubsection{Implementation Details.} \texttt{Training.} For latent diffusion experiments, we follow the standard training setup of \texttt{DiT}~\cite{peebles2023scalable} and \texttt{SiT}~\cite{ma2024sit}, training on ImageNet at $256 \times 256$ resolution with a batch size of $256$. For pixel space diffusion, we experiment with \texttt{DeCo-L/16} using a patch size of $16 \times 16$. For fair comparison with \texttt{ReDi}~\cite{kouzelis2025boosting}, we set $\lambda_z = 1$ and use a learnable projection with $8$ output channels. For pixel space diffusion, we set $\lambda_z = 0.1$ and use a learnable projection with $16$ output channels.
The projection layer is initialized with a random orthogonal matrix. Unless noted otherwise, we use the Feature Variance regularization loss with $\lambda_{\texttt{reg}} = 1.0$. For the \texttt{XL} experiments, which are trained for more iterations, we optimize the projection using a cosine decay scheduler; see the appendix for additional details.

\vspace{5pt}
\noindent \texttt{Sampling.} For latent diffusion experiments, we employ the
SDE Euler–Maruyama sampler. To ensure a fair comparison with \texttt{ReDi} the number of sampling steps is fixed at 250 across all experiments. For pixel diffusion experiments, we follow \texttt{DeCo} and use the Heun sampler and 50 sampling steps.

\vspace{5pt}
\noindent \texttt{Evaluation.}  To benchmark generative performance, we report Frechet Inception Distance (FID)~\cite{fid}, sFID\cite{sfid}, Inception Score~\cite{is}, Precision
(Pre.) and Recall (Rec.)~\cite{kynkaanniemi2019improved} using 50k samples and the ADM’s TensorFlow evaluation suite \cite{dhariwal2021diffusion}.

\subsection{Latent Space Diffusion}

\begin{table}[!h]
\centering
\small
\setlength{\tabcolsep}{4pt}
\begin{tabular}{c|lcccccc}
\toprule
\texttt{VE} & \texttt{Method} & \Th{FID$\downarrow$} & \Th{sFID$\downarrow$} & \Th{IS$\uparrow$} & \Th{Prec.$\uparrow$} & \Th{Rec.$\uparrow$} \\
\midrule
 & \texttt{ReDi}  & $30.9$ & $7.2$  & $49.2$ & $0.56$ & $0.57$ \\ 
\multirow{-2}{*}{\texttt{DINOv2}~\cite{oquab2023dinov2}} & \cellcolor{teal!10}\texttt{CoReDi} & \cellcolor{teal!10}$24.7$ & \cellcolor{teal!10}$6.7$  & \cellcolor{teal!10}$57.4$ & \cellcolor{teal!10}$0.60$ & \cellcolor{teal!10}$0.61$ \\
\arrayrulecolor{black}\midrule 
 & \texttt{ReDi}   & $38.2$ & $7.6$  & $37.4$ & $0.54$ & $0.58$ \\
\multirow{-2}{*}{\texttt{MOCOv3}~\cite{Chen2021AnES}} & \cellcolor{teal!10}\texttt{CoReDi} & \cellcolor{teal!10}$33.5$ & \cellcolor{teal!10}$6.5$  & \cellcolor{teal!10}$41.2$ & \cellcolor{teal!10}$0.55$ & \cellcolor{teal!10}$0.61$ \\
\arrayrulecolor{black}\midrule 
 & \texttt{ReDi}   & $36.2$ & $7.4$  & $41.5$ & $0.55$ & $0.59$ \\
\multirow{-2}{*}{\texttt{SigLIPv2}~\cite{tschannen2025siglip}} & \cellcolor{teal!10}\texttt{CoReDi} & \cellcolor{teal!10}$29.1$ & \cellcolor{teal!10}$6.5$  & \cellcolor{teal!10}$48.7$ & \cellcolor{teal!10}$0.58$ & \cellcolor{teal!10}$0.61$ \\
\arrayrulecolor{black}\midrule 
 & \texttt{ReDi}   & $40.3$ & $6.9$  & $34.1$ & $0.52$ & $0.58$ \\
\multirow{-2}{*}{\texttt{MAE}~\cite{he2022masked}} & \cellcolor{teal!10}\texttt{CoReDi} & \cellcolor{teal!10}$37.1$ & \cellcolor{teal!10}$6.6$  & \cellcolor{teal!10}$36.8$ & \cellcolor{teal!10}$0.52$ & \cellcolor{teal!10}$0.62$ \\
\bottomrule
\end{tabular}
\vspace{3pt}
\caption{\textbf{Representation Encoder Variation}. Results with different Visual Encoders (\texttt{VE}). All models at $200$K steps. \texttt{CoReDi} yields
consistent performance improvements across different choices for the visual encoder used for joint training compared to fixed ReDi projection used in \cite{kouzelis2025boosting}.}
\label{tab:ab_enc}
\end{table}

\subsubsection{CoReDi improves the generative modeling performance.}
To demonstrate the effectiveness of our approach, we compare \texttt{CoReDi} against \texttt{SiT}, \texttt{REPA}, and \texttt{ReDi} across different model scales and evaluation settings in \autoref{tab:fid_comparison} and \autoref{tab:sota_comparison}. Without classifier-free guidance, \texttt{CoReDi} consistently delivers better generative performance. For the \texttt{B/2} architecture, \texttt{CoReDi} reaches an FID of $16.4$ at $400$K iterations, substantially outperforming \texttt{ReDi} ($21.4$ FID) and \texttt{SiT} ($33.0$ FID) at the same training budget. For the larger \texttt{XL/2} architecture, \texttt{CoReDi} matches the best FID of $3.3$ achieved by \texttt{ReDi-XL/2}, while requiring only $2$M iterations instead of $4$M, and clearly outperforms both \texttt{REPA-XL/2} with $5.9$ FID and the converged \texttt{SiT-XL/2} baseline with  $8.3$ FID at $7$M iterations.

With classifier-free guidance, \texttt{CoReDi} further establishes a stronger trade-off between generation quality and training cost, as shown in \autoref{tab:sota_comparison}. \texttt{CoReDi} achieves an FID of $1.58$ after only $400$ epochs, improving over both \texttt{REPA} ($1.80$ at $800$ epochs) and \texttt{ReDi} ($1.72$ at $800$ epochs), while cutting the training steps by half.

\subsubsection{Generalization Across Visual Encoders.}
To further validate the generality of our approach, we evaluate \texttt{CoReDi} against \texttt{ReDi} across four different visual encoders in \autoref{tab:ab_enc}. \texttt{CoReDi} consistently improves over the fixed \texttt{PCA} projection of \texttt{ReDi} across all encoders, demonstrating that the benefits of our adaptive representation learning are not specific to any particular choice of visual encoder.  Notably, improvements are observed across all encoders, with the largest gains for \texttt{DINOv2} ($30.9 \rightarrow 24.7$ FID) and \texttt{SigLIPv2} ($36.2 \rightarrow 29.1$ FID), suggesting that the learned projection can extract more generative-relevant information for a variety of semantic features.

\subsection{Pixel Space Diffusion}

\subsubsection{Improved pixel space generation.}
For pixel space diffusion, we evaluate \texttt{CoReDi} on top of \texttt{DeCo-L/16} in \autoref{tab:main_res_pixel}. At $100$K iterations, \texttt{CoReDi-L/16} achieves a comparable FID of $31.5$ to \texttt{DeCo-L/16} at $200$K iterations, indicating a $\times2$ acceleration in convergence. With further training, the performance continues to improve, reaching an FID of $21.5$ at $200$K iterations. These results demonstrate that coevolving representations provide consistent benefits beyond latent-based generation, yielding notable performance improvements in the pixel diffusion setting as well.




\begin{table}[t]
\centering
\setlength{\tabcolsep}{4pt}
\begin{minipage}{0.45\linewidth}
    \vspace{22pt}
    \centering
    \begin{tabular}{lrrrrr}
    \toprule
    \Th{Model} & \Th{Iter.} & \Th{Param.} & \Th{FID$\downarrow$} \\
    \midrule
    \texttt{DeCo-L/16}*   & $100$K  & $426$M & $46.0$   \\
    \texttt{DeCo-L/16}   & $200$K  & $426$M & $31.3$   \\
    \rowcolor{teal!10} \texttt{CoReDi-L/16}   & $100$K & $426$M & $31.5$  \\
    \rowcolor{teal!10} \texttt{CoReDi-L/16}   & $200$K & $426$M & $21.5$  \\
    \bottomrule
    \end{tabular}
    \vspace{3pt}
    \caption{\textbf{Pixel Diffusion FID Comparisons}. FID scores on ImageNet $256\times256$ without Classifier-Free Guidance for \texttt{DeCo-L/16} and \texttt{CoReDi-L/16}. *Denotes our reproduction.}
    \label{tab:main_res_pixel}
\end{minipage}
\hfill
\begin{minipage}{0.52\linewidth}
    \centering
    \begin{tabular}{lrrrrrrr}
    \toprule
    $\lambda_z$ &  \Th{FID$\downarrow$} & \Th{sFID$\downarrow$} & \Th{IS$\uparrow$} & \Th{Prec.$\uparrow$} & \Th{Rec.$\uparrow$} \\
    \midrule
    $0.04$   & $34.4$ & $9.6$  & $43.0$ & $0.50$ & $0.63$ \\
    $0.08$   & $32.4$ & $9.1$  & $45.6$ & $0.51$ & $0.62$ \\
    \rowcolor{teal!10} $0.10$   & $31.5$ & $8.9$  & $47.0$ & $0.52$ & $0.63$ \\
    $0.14$   & $33.5$ & $10.2$  & $45.5$ & $0.50$ & $0.62$ \\
    \bottomrule
    \end{tabular}
    \vspace{3pt}
    \caption{\textbf{Pixel Space $\lambda_z$ Ablation} Results at $100$K steps for pixel \texttt{CoReDi-L/16} under different representation loss weights $\lambda_{\text{z}}$.}
    \label{tab:ab_reg_rep_weight}
\end{minipage}
\end{table}

\subsubsection{Representation Loss Weight in Pixel Space.}
In pixel space, images and semantic representations operate at very different dimensionalities, unlike the latent space setting where \texttt{VAE} latents and representations are of comparable scale. This discrepancy makes a naive transfer of the latent space hyperparameter $\lambda_z$ suboptimal. \autoref{tab:ab_reg_rep_weight} ablates the effect of $\lambda_z$ in the pixel space setting, showing that larger values that are effective in latent space diffusion lead to degraded performance here. Performance peaks at $\lambda_z = 0.1$, achieving an FID of $31.5$, which we adopt as our default for all pixel space experiments.

\subsection{Analysis}

\subsubsection{Necessity of Batch Normalization and Stop-Gradient.}
\autoref{tab:ab_sg_bn} validates the necessity of both batch normalization and stop-gradient in \texttt{CoReDi}. Removing the stop-gradient leads to a significant performance drop, with FID deteriorating from $24.7$ to $50.8$, as the projection trivially minimizes the representation loss by collapsing the target rather than learning a meaningful representation space. Removing batch normalization causes complete training collapse, with FID reaching $223.9$ and near-zero recall, confirming that the unbounded scale of the learned projection severely distorts the noise schedule and destabilizes training. These results confirm that both components are necessary for stable coevolution.

\newpage

\subsubsection{Effect of different Regularization Methods.} 

\begin{wraptable}{r}{0.38\textwidth}
\vspace{5pt}
\centering
\small
\setlength{\tabcolsep}{3pt}
\begin{tabular}{lrrr}
\toprule
\Th{Reg.} & \Th{FID$\downarrow$} & \Th{sFID$\downarrow$} & \Th{IS$\uparrow$} \\
\midrule
\rowcolor{gray!10} \texttt{ReDi} & $30.9$ & $7.2$ & $49.2$ \\
\texttt{-}     & $37.2$ & $7.2$ & $39.8$ \\
\texttt{Ortho} & $25.6$ & $6.8$ & $57.0$ \\
\texttt{Cov}   & $25.9$ & $7.0$ & $54.8$ \\
\rowcolor{teal!10} \texttt{VF} & $24.7$ & $6.7$ & $57.4$ \\
\bottomrule
\end{tabular}
\vspace{3pt}
\caption{\textbf{Effect of Regularization.} Results at $200$K steps for \texttt{CoReDi-B/2} under different regularization strategies: orthogonality (\texttt{Ortho}), covariance (\texttt{Cov}), feature variance (\texttt{VF}), and no regularization (\texttt{-}). The fixed \texttt{PCA} projection of \texttt{ReDi}~\cite{kouzelis2025boosting} is included as a reference (\textcolor{gray}{gray}).}
\label{tab:ab_reg}
\vspace{-20pt}
\end{wraptable}
In \autoref{tab:ab_reg} we present the effect of the different regularization strategies on generative performance. Without any regularization, the learned projection suffers from feature collapse, yielding an FID of $37.2$, worse than the static PCA projection of \texttt{ReDi}, confirming that naive joint optimization of the projection leads to degenerate solutions. This can also be qualitatively observed in \autoref{fig:vf_grid} for Feature Variance Loss. All three regularization strategies successfully prevent collapse and recover strong performance: orthogonality regularization achieves an FID of $25.6$, covariance regularization $25.9$ FID, and Feature Variance regularization $24.7$ FID. These results demonstrate that explicit regularization is a necessary ingredient for stable coevolution, and that Feature Variance regularization provides the best results among the strategies explored.

\begin{table*}[!t]
\centering
\begin{minipage}[t]{0.45\textwidth}
\centering
\setlength{\tabcolsep}{3pt}
\small
\scalebox{0.9}{
\begin{tabular}{lrrrrrrr}
\toprule
\Th{Reg.} &  \Th{FID$\downarrow$} & \Th{sFID$\downarrow$} & \Th{IS$\uparrow$} & \Th{Prec.$\uparrow$} & \Th{Rec.$\uparrow$} \\
\midrule
\rowcolor{teal!10}  \texttt{CoReDi}   & $24.7$ & $6.7$  & $57.4$ & $0.60$ & $0.61$ \\
w/o \texttt{SG}   & $50.8$ & $8.1$  & $29.5$ & $0.49$ & $0.53$ \\
w/o \texttt{BN}   & $223.9$ & $150.6$  & $3.6$ & $0.06$ & $0.01$ \\
\bottomrule
\end{tabular}
}
\vspace{3pt}
\captionsetup{type=table, margin={5pt,-10pt}}
\caption{\textbf{Stabilization Methods.} Results at $200$K steps for latent \texttt{CoReDi-B/2} without Stop-Gradient (\texttt{SG}) and without Batch Normalization (\texttt{BN}).}
\label{tab:ab_sg_bn}
\end{minipage}
\hfill
\begin{minipage}[t]{0.50\textwidth}
\centering
\setlength{\tabcolsep}{3pt}
\small
\scalebox{0.9}{
\begin{tabular}{lccccccc}
\toprule
$\lambda_{\text{reg}}$ &  \Th{FID$\downarrow$} & \Th{sFID$\downarrow$} & \Th{IS$\uparrow$} & \Th{Prec.$\uparrow$} & \Th{Rec.$\uparrow$} \\
\midrule
$0.5$   & $25.9$ & $6.6$  & $55.2$ & $0.59$ & $0.60$ \\
$1.0$   & $24.7$ & $6.7$  & $57.4$ & $0.60$ & $0.61$ \\
\rowcolor{teal!10}  $1.5$   & $24.3$ & $6.6$  & $58.9$ & $0.61$ & $0.61$ \\
\bottomrule
\end{tabular}
}
\vspace{3pt}
\captionsetup{type=table, margin={10pt,5pt}}
\caption{\textbf{Regularization Weight.} Results for \texttt{CoReDi-B/2} at 200K under different regularization weights $\lambda_{\text{reg}}$ for Feature Variance Loss.}
\label{tab:ab_variance_feats}
\end{minipage}
\end{table*}

\subsubsection{Effect of Feature Variance Regularization Weight.}
In \autoref{tab:ab_variance_feats} we ablate the effect of the variance regularization weight $\lambda_{\text{reg}}$. All three values yield strong performance, demonstrating robustness to this hyperparameter. Performance improves slightly with larger weights, with $\lambda_{\text{reg}} = 1.5$ achieving the best FID of $24.3$.

\subsection{Coevolving Representations Develop Spatial Structure}

Recent work has shown that the spatial structure of visual representations, rather than their global semantic quality, is the primary driver of generation performance in representation-guided diffusion models~\cite{singh2025matters}. Motivated by this finding, we investigate whether the coevolving representations in \texttt{CoReDi} develop stronger spatial structure over the course of training. We measure this using three spatial self-similarity metrics proposed in \cite{singh2025matters}: \texttt{LDS}~\cite{lds}, which measures the contrast between the average similarity of nearby versus distant patch pairs, \texttt{CDS}~\cite{singh2025matters}, which captures how quickly patch similarity decays with spatial distance, and \texttt{RMSC}~\cite{singh2025matters}, which measures the overall spatial diversity of patch features. (See Appendix for more details).

\definecolor{dustydustypurple}{HTML}{7B6FA0} 
\begin{figure}[t]
  \centering
  \begin{minipage}[t]{0.31\linewidth}
    \centering
  \begin{tikzpicture}
  \begin{axis}[
    width=1.2\linewidth,
    height=0.9\linewidth,
    xmin=-10, xmax=210,
    ymin=0.19, ymax=0.46,
    tick label style={font=\scriptsize},
    label style={font=\scriptsize},
    xlabel={Training Steps},
    title={\textbf{LDS}},
    xtick={0,50,100,150,200},
    xticklabels={0K,50k,100k,150k,200k},
    grid=major,
    major grid style={gray!20},
  ]
    \addplot[color=teal, mark=*] coordinates {
      (0, 0.25)
      (50, 0.4053892475366592)
      (100, 0.4260256424546)
      (150, 0.4416864302)
      (200, 0.4494220411)
    };
    \addplot[color=teal, dashed, thick] coordinates {
      (-10, 0.22)
      (220, 0.22)
    };
  \end{axis}
\end{tikzpicture}
  \end{minipage}
  \begin{minipage}[t]{0.31\linewidth}
    \centering
\begin{tikzpicture}
  \begin{axis}[
    width=1.2\linewidth,
    height=0.9\linewidth,
    xmin=-10, xmax=210,
    ymin=0.030, ymax=0.065,
    tick label style={font=\scriptsize},
    label style={font=\scriptsize},
    xlabel={Training Steps},
    title={\textbf{CDS}},
    xtick={0,50,100,150,200},
    xticklabels={0K,50k,100k,150k,200k},
    grid=major,
    major grid style={gray!20},
  ]
    \addplot[color=teal, mark=*] coordinates {
      (0, 0.04)
      (50, 0.057911827601492404)
      (100, 0.06049077287316322)
      (150, 0.06247813131660223)
      (200, 0.06356985963881016)
    };
    \addplot[color=teal, dashed, thick] coordinates {
      (-10, 0.033)
      (220, 0.033)
    };
  \end{axis}
\end{tikzpicture}
  \end{minipage}\hspace{-0.5em}
  \begin{minipage}[t]{0.31\linewidth}
    \centering
\begin{tikzpicture}
  \begin{axis}[
    width=1.2\linewidth,
    height=0.9\linewidth,
    xmin=-10, xmax=210,
    ymin=0.72, ymax=0.8,
    tick label style={font=\scriptsize},
    label style={font=\scriptsize},
    xlabel={Training Steps},
    title={\textbf{RMSC}},
    xtick={0,50,100,150,200},
    xticklabels={0K,50k,100k,150k,200k},
    grid=major,
    major grid style={gray!20},
  ]
    \addplot[color=teal, mark=*] coordinates {
      (0, 0.7523464)
      (50, 0.7442838430404664)
      (100, 0.7683448958396911)
      (150, 0.7846799111366272)
      (200, 0.7927114498615265)
    };
    \addplot[color=teal, dashed, thick] coordinates {
      (-10, 0.73)
      (220, 0.73)
    };
  \end{axis}
\end{tikzpicture}
  \end{minipage}
  \\[4pt]
  \begin{minipage}[t]{0.31\linewidth}
    \centering
  \begin{tikzpicture}
  \begin{axis}[
    width=1.2\linewidth,
    height=0.9\linewidth,
    xmin=-10, xmax=210,
    ymin=0.23, ymax=0.33,
    tick label style={font=\scriptsize},
    label style={font=\scriptsize},
    xlabel={Training Steps},
    xtick={0,50,100,150,200},
    xticklabels={0K,50k,100k,150k,200k},
    grid=major,
    major grid style={gray!20},
  ]
    \addplot[color=dustypurple, mark=*] coordinates {
      (0, 0.245750357598126)
      (50, 0.26465750992298126)
      (100, 0.28842667743563655)
      (150, 0.31039744436740874)
      (200, 0.32178458631038664)
    };
    \addplot[color=dustypurple, dashed, thick] coordinates {
      (-10, 0.25656659174710512)
      (220, 0.25656659174710512)
    };
  \end{axis}
\end{tikzpicture}
  \end{minipage}
  \begin{minipage}[t]{0.31\linewidth}
    \centering
\begin{tikzpicture}
  \begin{axis}[
    width=1.2\linewidth,
    height=0.9\linewidth,
    xmin=-10, xmax=210,
    ymin=0.028, ymax=0.045,
    tick label style={font=\scriptsize},
    label style={font=\scriptsize},
    xlabel={Training Steps},
    xtick={0,50,100,150,200},
    xticklabels={0K,50k,100k,150k,200k},
    grid=major,
    major grid style={gray!20},
  ]
    \addplot[color=dustypurple, mark=*] coordinates {
      (0, 0.03)
      (50, 0.03537213023751974)
      (100, 0.03890821807086468)
      (150, 0.04246396418660879)
      (200, 0.0444732765853405)
    };
    \addplot[color=dustypurple, dashed, thick] coordinates {
      (-10, 0.029)
      (220, 0.029)
    };
  \end{axis}
\end{tikzpicture}
  \end{minipage}\hspace{-0.5em}
  \begin{minipage}[t]{0.31\linewidth}
    \centering
\begin{tikzpicture}
  \begin{axis}[
    width=1.2\linewidth,
    height=0.9\linewidth,
    xmin=-10, xmax=210,
    ymin=0.61, ymax=0.77,
    tick label style={font=\scriptsize},
    label style={font=\scriptsize},
    xlabel={Training Steps},
    xtick={0,50,100,150,200},
    xticklabels={0K,50k,100k,150k,200k},
    grid=major,
    major grid style={gray!20},
  ]
    \addplot[color=dustypurple, mark=*] coordinates {
      (0,0.62)
      (50, 0.66)
      (100, 0.70)
      (150, 0.73)
      (200, 0.75)
    };
    \addplot[color=dustypurple, dashed, thick] coordinates {
      (-10, 0.62)
      (220, 0.62)
    };
  \end{axis}
\end{tikzpicture}
  \end{minipage}
  \\[4pt]
  \vspace{-4pt}
  \caption{ \textbf{Spatial structure of coevolving representations during training} for \texttt{CoReDi} with \textcolor{teal}{\texttt{DINOv2}} and \textcolor{dustypurple}{\texttt{MOCOv3}} as measured by \texttt{LDS}, \texttt{CDS}, and \texttt{RMSC}. All three metrics improve consistently as training progresses. Dashed horizontal lines indicate the fixed PCA projections used in \texttt{ReDi} \cite{kouzelis2025boosting}.}
  \label{fig:spatial_sim}
\end{figure}
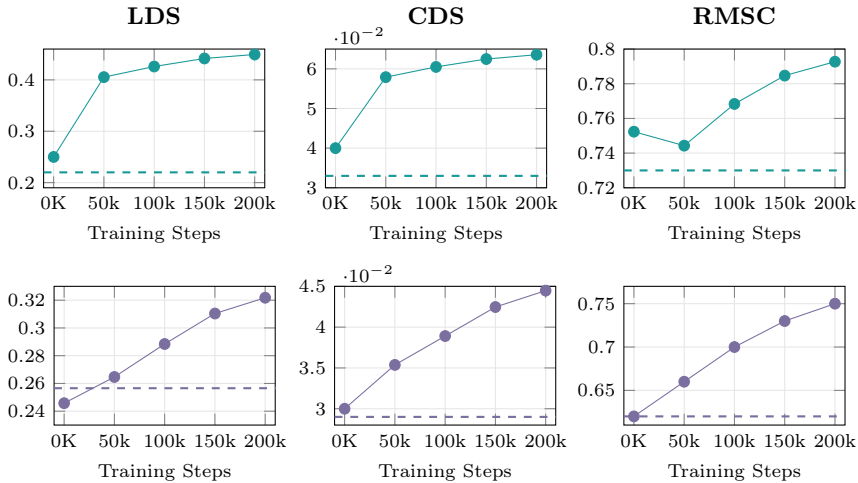

In \autoref{fig:spatial_sim}, we illustrate the evolution of all three metrics throughout \texttt{CoReDi} training. We observe that all three improve consistently as training progresses, suggesting that the adaptive projection naturally evolves toward representations with stronger spatial organization. Furthermore, the learned projections achieve higher spatial structure scores than the fixed \texttt{PCA} projection used in \texttt{ReDi}, suggesting that the coevolving representation space captures richer spatial information than a static linear projection. This suggests a potential explanation for the generative improvements observed with \texttt{CoReDi}: the joint optimization of the projection with the generative objective encourages the learned representation space to develop spatial structure that is more beneficial for image synthesis.

\section{Conclusion}
We introduced \texttt{CoReDi}, a framework for joint image-feature diffusion in which the semantic representation space coevolves with the generative model during training. Unlike prior work that relies on fixed, predetermined representations to assist generation, \texttt{CoReDi} learns an adaptive projection of the representation space jointly with the diffusion objective, allowing the representation space to develop structure that is directly beneficial for image synthesis. Through systematic analysis, we identified three necessary ingredients for stable coevolution — stop-gradient stabilization, batch normalization, and explicit regularization against feature collapse — and demonstrated empirically that all three are essential. We further showed that the coevolving representations develop stronger spatial structure over the course of training, providing a potential explanation for the observed generative improvements. Finally, we demonstrated that \texttt{CoReDi} extends naturally beyond \texttt{VAE} latent spaces to pixel-space diffusion, yielding consistent improvements across both settings. We hope that this work motivates further exploration of adaptive representation spaces as a tool for improving generative modeling.

\section*{Acknowledgements}
This work has been partially supported by project MIS 5154714 of the National Recovery and Resilience Plan Greece 2.0 funded by the European Union under the NextGenerationEU Program.
Hardware resources were granted with the support of GRNET. Also, this work was partially conducted using EuroHPC resources (Project ID e-dev-2026d01-087) and HPC resources from GENCI-IDRIS (Grant AD011016639).


\bibliographystyle{splncs04}
\bibliography{main}
\newpage

\newpage

{\Huge \textbf{Appendix}}

\appendix

\section{Additional Results and Ablations}




\subsubsection{Detailed Quantitative Comparison.}
\definecolor{lightgraycustom}{gray}{0.9}
\definecolor{midgraycustom}{gray}{0.70}

\begin{table}[!h]
\centering
\setlength{\tabcolsep}{4pt}
\begin{tabular}{lrrrrrr}
\toprule
\Th{Model} & \Th{\#Iters.} & \Th{FID$\downarrow$} & \Th{sFID$\downarrow$} & \Th{IS$\uparrow$} & \Th{Prec.$\uparrow$} & \Th{Rec.$\uparrow$} \\
\midrule
\rowcolor{midgraycustom} \texttt{SiT-XL/2}~\cite{ma2024sit} & $7\text{M}$ & $8.3$ & $6.3$ & $131.7$ & $0.68$ & $0.67$ \\
\midrule
 \texttt{REPA-XL/2}~\cite{yu2025repa} & $50\text{K}$ & $52.3$ & $31.2$ & $24.3$ & $0.45$ & $0.53$ \\
 \texttt{ReDi-XL/2}~\cite{kouzelis2025boosting} & $50\text{K}$ & $56.1$ & $18.9$ & $23.8$ & $0.44$ & $0.47$ \\
\rowcolor{teal!10} \texttt{CoReDi-XL/2} & $50\text{K}$ & $40.9$ & $9.74$ & $32.1$ & $0.53$ & $0.56$ \\
\midrule
 \texttt{REPA-XL/2}~\cite{yu2025repa} & $100\text{K}$ & $19.4$ & $6.1$ & $67.4$ & $0.64$ & $0.61$ \\
\texttt{ReDi-XL/2}~\cite{kouzelis2025boosting} & $100\text{K}$ & $23.1$ & $5.9$ & $61.5$ & $0.64$ & $0.57$ \\
\rowcolor{teal!10} \texttt{CoReDi-XL/2} & $100\text{K}$ & $19.0$ & $5.6$ & $69.3$ & $0.65$ & $0.59$ \\
\midrule
 \texttt{REPA-XL/2}~\cite{yu2025repa} & $200\text{K}$ & $11.1$ & $5.0$ & $100.4$ & $0.69$ & $0.64$ \\
 \texttt{ReDi-XL/2}~\cite{kouzelis2025boosting} & $200\text{K}$ & $12.6$ & $5.7$ & $97.3$ & $0.69$ & $0.61$ \\
\rowcolor{teal!10} \texttt{CoReDi-XL/2} & $200\text{K}$ & $9.2$ & $4.7$ & $110.0$ & $0.71$ & $0.62$ \\
\midrule
 \texttt{REPA-XL/2}~\cite{yu2025repa} & $400\text{K}$ & $7.9$ & $5.1$ & $122.6$ & $0.70$ & $0.65$ \\
 \texttt{ReDi-XL/2}~\cite{kouzelis2025boosting} & $400\text{K}$ & $7.5$ & $5.1$ & $129.5$ & $0.72$ & $0.62$ \\
\rowcolor{teal!10} \texttt{CoReDi-XL/2} & $400\text{K}$ & $6.1$ & $4.6$ & $136.1$ & $0.73$ & $0.64$ \\
\midrule
\texttt{REPA-XL/2}~\cite{yu2025repa} & $4\text{M}$ & $5.9$ & $5.7$ & $157.8$ & $0.70$ & $0.69$ \\
 \texttt{ReDi-XL/2}~\cite{kouzelis2025boosting} & $4\text{M}$ & $3.3$ & $4.8$ & $188.9$ & $0.74$ & $0.68$ \\
\rowcolor{teal!10} \texttt{CoReDi-XL/2} & $2\text{M}$ & $3.3$ & $4.4$ & $176.8$ & $0.74$ & $0.66$ \\
\bottomrule
\end{tabular}
\vspace{3pt}
\caption{\textbf{Detailed evaluation} for \texttt{SiT-XL/2}, \texttt{CoReDi-XL/2}, \texttt{ReDi-XL/2}, and \texttt{REPA-XL/2}. All results are reported without Classifier-Free Guidance.}
\label{tab:details_coredi}
\end{table}
In Table~\ref{tab:details_coredi}, we present detailed results for \texttt{CoReDi} alongside \texttt{REPA} and \texttt{ReDi}. We observe that \texttt{CoReDi} converges significantly faster than both baselines and matches \texttt{ReDi}'s converged generative performance at 4\text{M} iterations with only 2\text{M} iterations.

\subsubsection{VAE-only Classifier-Free Guidance.}
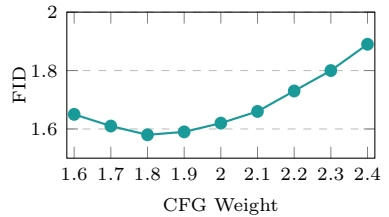
\begin{wrapfigure}{r}{0.42\textwidth}
    \vspace{-30pt}
    \centering
    \begin{tikzpicture}
    \begin{axis}[
        xlabel={CFG Weight},
        ylabel={FID},
        xmin=1.58, xmax=2.42,
        ymin=1.5, ymax=2.0,
        xtick={1.6,1.7,1.8,1.9,2.0,2.1,2.2,2.3,2.4},
        ymajorgrids=true,
        grid style=dashed,
        width=5.7cm,
        height=3.5cm,
        label style={font=\scriptsize},
        tick label style={font=\scriptsize},
        ylabel style={font=\scriptsize, yshift=-5pt},
    ]
    \addplot[
        color=teal,
        mark=*,
        thick,
    ]
    coordinates {
        (1.6, 1.65)
        (1.7, 1.61)
        (1.8, 1.58)
        (1.9, 1.59)
        (2.0, 1.62)
        (2.1, 1.66)
        (2.2, 1.73)
        (2.3, 1.80)
        (2.4, 1.89)
    };
    \end{axis}
    \end{tikzpicture}
    \vspace{-20pt}
    \caption{FID score as a function of CFG weight.}
    \label{fig:cfg_weight}
    \vspace{-20pt}
\end{wrapfigure}
Following \texttt{ReDi}~\cite{kouzelis2025boosting}, we apply Classifier-Free Guidance exclusively to the VAE latents rather than across both the image latents and the features, as this strategy consistently yields superior generation quality and greater robustness to CFG weight variations (see Section 4.4 in \cite{kouzelis2025boosting} for more details). In our ablation as presented in \autoref{fig:cfg_weight}, we find that a CFG weight of $1.8$ achieves optimal performance.



\subsubsection{Spatial Structure of Coevolving Representations in Pixel Diffusion.} 
\begin{figure}[h]
  \centering
  \begin{minipage}[t]{0.31\linewidth}
    \centering
  \begin{tikzpicture}
  \begin{axis}[
    width=1.15\linewidth,
    height=0.9\linewidth,
    xmin=-10, xmax=210,
    ymin=0.19, ymax=0.46,
    tick label style={font=\scriptsize},
    label style={font=\scriptsize},
    xlabel={Training Steps},
    ylabel={\textbf{LDS}},
    xtick={0,50,100,150,200},
    xticklabels={0K,50k,100k,150k,200k},
    grid=major,
    major grid style={gray!20},
  ]
    \addplot[color=teal, mark=*] coordinates {
      (0, 0.248)
      (50, 0.3828)
      (100, 0.4181)
      (150, 0.4402)
      (200, 0.4479)
    };
    \addplot[color=teal, dashed, thick] coordinates {
      (-10, 0.22)
      (220, 0.22)
    };
  \end{axis}
\end{tikzpicture}
  \end{minipage}
  \begin{minipage}[t]{0.31\linewidth}
    \centering
\begin{tikzpicture}
  \begin{axis}[
    width=1.1\linewidth,
    height=0.9\linewidth,
    xmin=-10, xmax=210,
    ymin=0.030, ymax=0.065,
    tick label style={font=\scriptsize},
    label style={font=\scriptsize},
    xlabel={Training Steps},
    ylabel={\textbf{CDS}},
    xtick={0,50,100,150,200},
    xticklabels={0K,50k,100k,150k,200k},
    grid=major,
    major grid style={gray!20},
  ]
    \addplot[color=teal, mark=*] coordinates {
      (0, 0.0395)
      (50, 0.0474)
      (100, 0.0559)
      (150, 0.0621)
      (200, 0.0632)
    };
    \addplot[color=teal, dashed, thick] coordinates {
      (-10, 0.033)
      (220, 0.033)
    };
  \end{axis}
\end{tikzpicture}
  \end{minipage}\hspace{-0.5em}
  \begin{minipage}[t]{0.31\linewidth}
    \centering
\begin{tikzpicture}
  \begin{axis}[
    width=1.1\linewidth,
    height=0.9\linewidth,
    xmin=-10, xmax=210,
    ymin=0.72, ymax=0.8,
    tick label style={font=\scriptsize},
    label style={font=\scriptsize},
    xlabel={Training Steps},
    ylabel={\textbf{RMSC}},
    xtick={0,50,100,150,200},
    xticklabels={0K,50k,100k,150k,200k},
    grid=major,
    major grid style={gray!20},
  ]
    \addplot[color=teal, mark=*] coordinates {
      (0, 0.7511)
      (50, 0.7553)
      (100, 0.7675)
      (150, 0.7836)
      (200, 0.7918)
    };
    \addplot[color=teal, dashed, thick] coordinates {
      (-10, 0.73)
      (220, 0.73)
    };
  \end{axis}
\end{tikzpicture}
  \end{minipage}

  \vspace{-4pt}
  \caption{ \textbf{Spatial structure of coevolving representations in pixel space} for \texttt{CoReDi-L/16} with \textcolor{teal}{\texttt{DINOv2}} as measured by \texttt{LDS}, \texttt{CDS}, and \texttt{RMSC}. All three metrics improve consistently as training progresses. Dashed horizontal lines indicate the fixed PCA projections.}
  \label{fig:spatial_sim_appendix}
\end{figure}
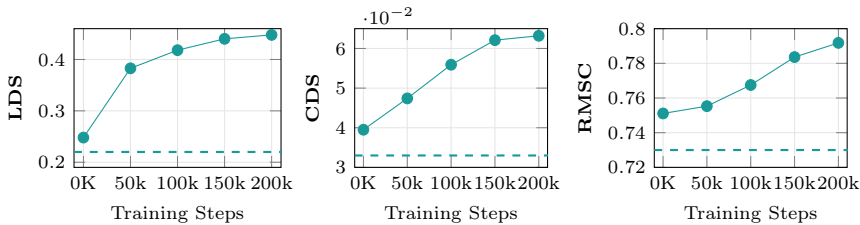

\autoref{fig:spatial_sim_appendix} shows the evolution of spatial structure metrics throughout \texttt{CoReDi} training in pixel space. All three metrics — \texttt{LDS}, \texttt{CDS}, and \texttt{RMSC} — improve consistently as training progresses. This validates our observation that coevolving representations develop increasingly structured spatial organization during training, and further demonstrates that this phenomenon is not specific to latent space diffusion but holds in the pixel space setting as well.

\section{Additional Implementation Details}

\subsection{Architecture settings}
\subsubsection{Latent Space Diffusion.}
We follow the \texttt{SiT} configurations from \cite{ma2024sit}. \texttt{SiT-B/2} ($130$M parameters) uses $12$ transformer blocks with embedding dimension $768$ and $12$ attention heads. \texttt{SiT-XL/2} ($675$M parameters) uses $28$ blocks with embedding dimension $1152$ and $16$ heads. In all latent diffusion experiments, images are encoded with \texttt{SD-VAE-FT-EMA}, and we use a $2\times2$ patch size. 

We observe that applying a cosine decay schedule to the projection yields more stable optimization over a longer training horizon and results in better generative performance. In particular, for the \texttt{XL} experiments, we use a cosine decay schedule that reduces the projection learning rate to 0 by $400$k iterations. We emphasize that this schedule is applied only to the learnable projection and not to the backbone \texttt{DiT}, which is trained with the standard constant learning rate $1e-4$.

\subsubsection{Pixel Space Diffusion.}
We follow the \texttt{DeCo} configuration from \cite{ma2025deco}. \texttt{DeCo-L/16} uses $22$ transformer blocks in the encoder and $3$ MLP blocks in the pixel decoder, with embedding dimension $1024$ and $16$ attention heads. In pixel-space experiments, we set $\lambda_z = 0.1$ and use a learnable projection to $16$ channels. 

\subsection{Optimization Settings}

We optimize all models using AdamW~\cite{kingma2014adam} 
with a constant learning rate of $1 \times 10^{-4}$, momentum parameters $(\beta_1, \beta_2) = (0.9, 0.999)$, 
and a batch size of 256. To accelerate training for latent diffusion we pre-compute the image latents. 
Full optimization details are provided in Table~\ref{appendix:tab_config}.

\label{app:implem_details}
\begin{table}[t]
\centering
\begin{tabular}{l |ccc}
 & \textbf{Pixel} & \multicolumn{2}{c}{\textbf{Latent}} \\
\rowcolor[gray]{0.9}\multicolumn{4}{l}{\textbf{CoReDi}} \\
$\lambda_z$ & 0.1 & 1 & 1 \\
$\lambda_{reg}$ & 1 & 1 & 1 \\
Proj. Channels & 16 & 8 & 8 \\
\rowcolor[gray]{0.9}\multicolumn{4}{l}{\textbf{Architecture}} \\
Backbone & \texttt{DeCo-L} & \texttt{SiT-B} & \texttt{SiT-XL} \\
DiT Depth & 22 & 12 & 28 \\
Hidden Dim. & 1024 & 768 & 1152 \\
Heads & 16 & 12 & 16 \\
Params & 426M & 130M & 675M \\
Decoder Depth & 3 & - & - \\
Decoder Hidden Dim. & 32 & - & - \\
Patch Size & 16 & 2 & 2 \\
\rowcolor[gray]{0.9}\multicolumn{4}{l}{\textbf{Optimization}} \\
Optimizer & AdamW & AdamW & AdamW \\
Batch Size & 256 & 256 & 256 \\
Learning Rate & 1e-4 & 1e-4 & 1e-4 \\
LR Schedule  & constant & constant & constant \\
LR Schedule Proj.  & constant & constant & cos. decay \\
Weight Decay & 0 & 0 & 0 \\
EMA Decay & 0.9999 & 0.9999 & 0.9999 \\
Time Sampler & $\text{logit normal}$ & $\text{uniform}$ & $\text{uniform}$ \\
\rowcolor[gray]{0.9}\multicolumn{4}{l}{\textbf{Sampling}} \\
Steps & 50 & 250 & 250 \\
CFG Scale & - & - & 1.8 \\
\end{tabular}
\vspace{10pt}
\caption{{Configurations of pixel space and latent space experiments.}}
\label{appendix:tab_config}
\end{table}

\section{Evaluation Metrics}

\subsubsection{Generative Performance Evaluation Metrics.} We follow the evaluation setup of ADM~\cite{dhariwal2021diffusion}, generating $50$K samples and computing metrics using the official TensorFlow evaluation toolkit~\cite{dhariwal2021diffusion}. All evaluations are conducted on NVIDIA A100 GPUs. We report the following metrics:

\begin{itemize}
    \item \noindent \textbf{FID}~\cite{fid} measures the distance between the feature distributions of real and generated images, computed using \texttt{Inception-v3} under a multivariate Gaussian assumption.

   \item \noindent \textbf{sFID}~\cite{sfid} computes FID using spatial feature maps from intermediate \\ \texttt{Inception-v3} layers, better capturing the spatial structure of generated images.

    \item \noindent \textbf{IS}~\cite{is} evaluates generated images using \texttt{Inception-v3}, assigning higher scores to outputs that are both classifiable with high confidence and diverse across categories.

    \item \noindent \textbf{Precision \& Recall}~\cite{kynkaanniemi2019improved} measure realism and diversity in feature space. Precision reflects the fraction of generated images that appear realistic, while recall measures coverage of the real data distribution.

\end{itemize}

\subsection{Spatial Structure Evaluation Metrics}

In this section, we briefly describe each spatial self-similarity metric used to analyze the coevolving representations in \texttt{CoReDi}, following the definitions introduced in \cite{singh2025matters}.

\noindent \textbf{Local vs.\ Distant Similarity (LDS).} LDS measures the average similarity contrast between spatially close and distant patch pairs:
\[
\text{LDS}(\mathbf{X}) = \mathbb{E}[K_\mathbf{X}(t,t') \mid d(t,t') < r_{\text{near}}] - \mathbb{E}[K_\mathbf{X}(t,t') \mid d(t,t') \geq r_{\text{far}}],
\]
where $K_\mathbf{X}$ is the cosine similarity between patch tokens and $d(\cdot,\cdot)$ is the Manhattan distance. Larger values indicate stronger spatial organization, where nearby patches are more similar to each other than distant ones.

\noindent \textbf{Correlation Decay Slope (CDS).} CDS measures how quickly patch similarity decays with spatial distance. Given the spatial correlogram $g_\mathbf{X}(\delta) = \mathbb{E}[K_\mathbf{X}(t,t') \mid d(t,t') = \delta]$, fit a least-squares line $\hat{g}_\mathbf{X}(\delta) \approx \alpha + \beta\delta$ and define:
\[
\text{CDS}(\mathbf{X}) = -\hat{\beta},
\]
where $\hat{\beta}$ is the fitted slope. Larger values indicate faster similarity decay with distance, reflecting stronger spatial organization.

\noindent \textbf{RMS Spatial Contrast (RMSC).} RMSC measures the spatial diversity of patch token representations. Given normalized patch features $\hat{\mathbf{x}}_t = \mathbf{x}_t / \|\mathbf{x}_t\|_2$, it is defined as:
\[
\text{RMSC}(\mathbf{X}) = \sqrt{\frac{1}{T} \sum_{t=1}^{T} \|\hat{\mathbf{x}}_t - \bar{\mathbf{x}}\|_2^2},
\]
where $\bar{\mathbf{x}} = \frac{1}{T}\sum_{t=1}^{T} \hat{\mathbf{x}}_t$ is the mean normalized feature. Higher values indicate greater spatial diversity, reflecting preserved spatial structure, while lower values indicate more uniform, spatially uninformative representations.

\section{Additional Qualitative Results}

\begin{figure*}[h]
    \centering
    \setlength{\tabcolsep}{1pt} 
    \renewcommand{\arraystretch}{1.0} 
    \begin{tabular}{lcccccc}
        \rotatebox{90}{%
          \multirow{2}{*}{$\;\;\;$\texttt{Image}}%
        } & 
        \includegraphics[width=0.14\linewidth]{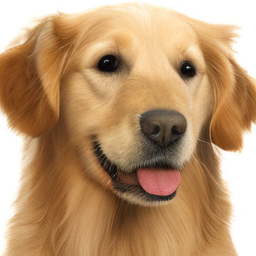} &
        \includegraphics[width=0.14\linewidth]{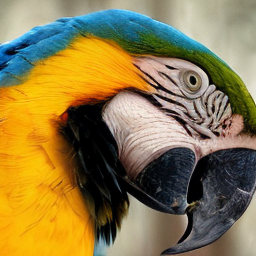} &
        \includegraphics[width=0.14\linewidth]{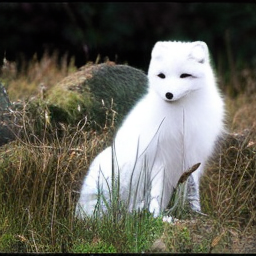} &
        \includegraphics[width=0.14\linewidth]{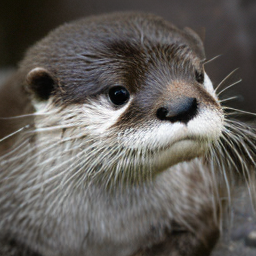} &
        \includegraphics[width=0.14\linewidth]{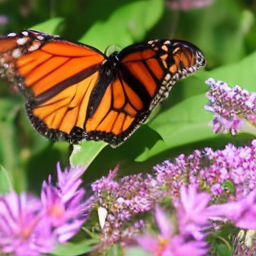} &
        \includegraphics[width=0.14\linewidth]{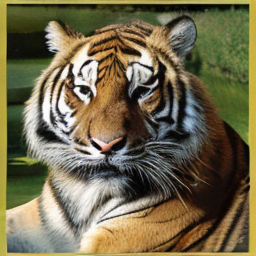} 

        \\

        \rotatebox{90}{%
          \begin{tabular}{@{}l@{}}
            \texttt{$\;\;$Feature}
          \end{tabular}%
        } & 
        \includegraphics[width=0.14\linewidth]{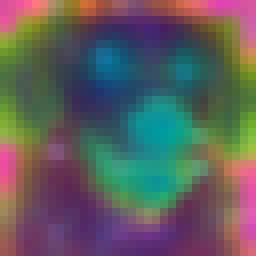} &
        \includegraphics[width=0.14\linewidth]{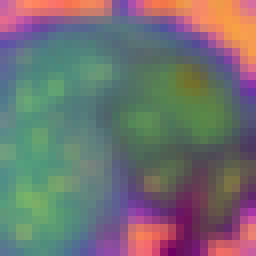} &
        \includegraphics[width=0.14\linewidth]{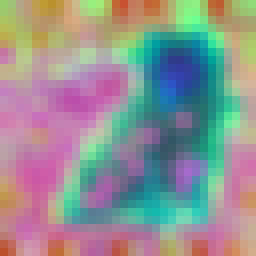} &
        \includegraphics[width=0.14\linewidth]{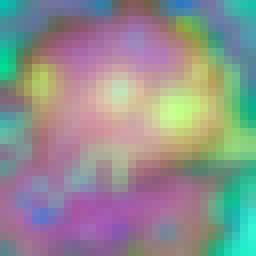} &
        \includegraphics[width=0.14\linewidth]{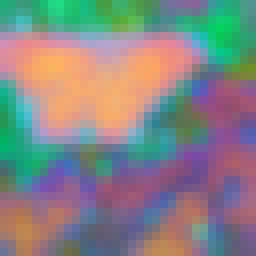} &
        \includegraphics[width=0.14\linewidth]{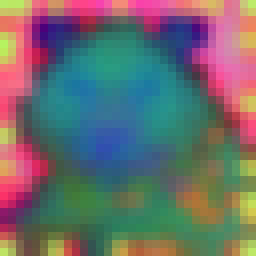} \\
    \end{tabular}

    \caption{\textbf{Selected samples} from our \texttt{CoReDi-XL/2} trained for $1$M steps on ImageNet $256\times256$. Images and visual representations are jointly generated by our model. We use Classifier-Free Guidance with $w = 4.0$.}
    \label{fig:qual}
    \vspace{-10pt}
\end{figure*}

We provide qualitative results of both generated images and visual representations in \autoref{fig:qual}. Further, in \autoref{fig:pca_vis_appendix}, we visually compare our learned representations with static PCA for all visual encoders examined in the paper.
\label{app:figures}

\definecolor{warmgold}{HTML}{C9973A}
\definecolor{coralrose}{HTML}{C96A6A}
\definecolor{sage}{HTML}{7A9E7E}        
\definecolor{dustyrose}{HTML}{C28B8B}   
\definecolor{slate}{HTML}{6B7FA3}       
\definecolor{amber}{HTML}{D4943A}       
\definecolor{mauve}{HTML}{9B7FA3}       
\definecolor{terracotta}{HTML}{C27A5A}  
\definecolor{seafoam}{HTML}{6AADA8}     
\definecolor{lavender}{HTML}{8A7DB8}    
\definecolor{sienna}{HTML}{B87355}      
\definecolor{steelrose}{HTML}{B87A8A}   

\begin{figure}[t]
\centering
\setlength{\tabcolsep}{2pt}
\begin{tabular}{c|cc|cc|cc|cc}
    \multicolumn{1}{c}{\texttt{\textbf{Image}}}
    & \multicolumn{1}{c}{\texttt{\textbf{PCA}}}
    & \multicolumn{1}{c}{\texttt{\textbf{CoReDi}}}
    & \multicolumn{1}{c}{\texttt{\textbf{PCA}}}
    & \multicolumn{1}{c}{\texttt{\textbf{CoReDi}}}
    & \multicolumn{1}{c}{\texttt{\textbf{PCA}}}
    & \multicolumn{1}{c}{\texttt{\textbf{CoReDi}}}
    & \multicolumn{1}{c}{\texttt{\textbf{PCA}}}
    & \multicolumn{1}{c}{\texttt{\textbf{CoReDi}}} \\[4pt]
    \includegraphics[width=0.10\linewidth]{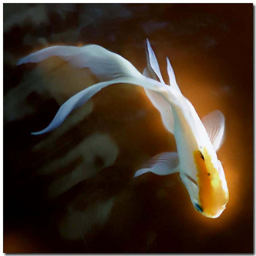} &
    \includegraphics[width=0.10\linewidth]{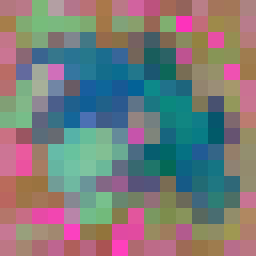} &
    \includegraphics[width=0.10\linewidth]{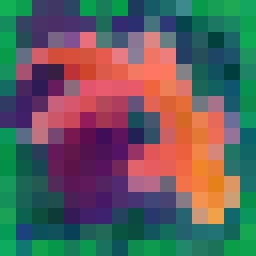} &
    \includegraphics[width=0.10\linewidth]{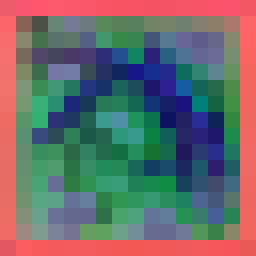} &
    \includegraphics[width=0.10\linewidth]{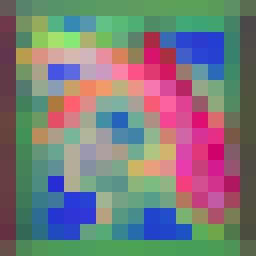} &
    \includegraphics[width=0.10\linewidth]{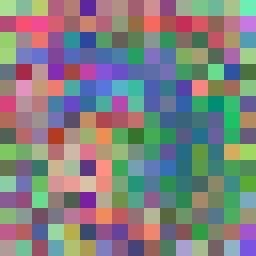} &
    \includegraphics[width=0.10\linewidth]{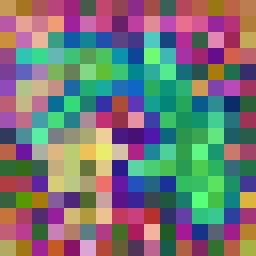} &
    \includegraphics[width=0.10\linewidth]{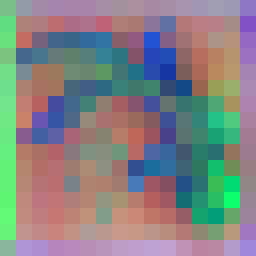} &
    \includegraphics[width=0.10\linewidth]{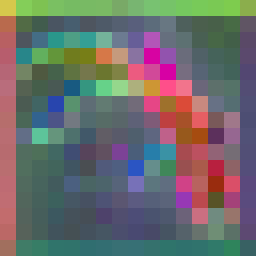} \\[2pt]
    \includegraphics[width=0.10\linewidth]{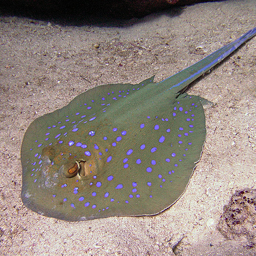} &
    \includegraphics[width=0.10\linewidth]{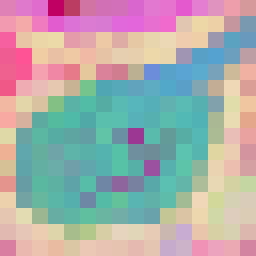} &
    \includegraphics[width=0.10\linewidth]{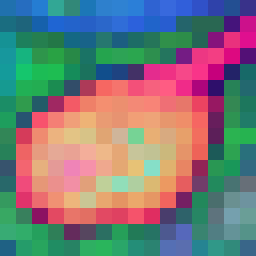} &
    \includegraphics[width=0.10\linewidth]{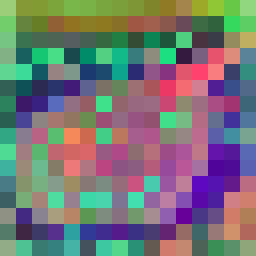} &
    \includegraphics[width=0.10\linewidth]{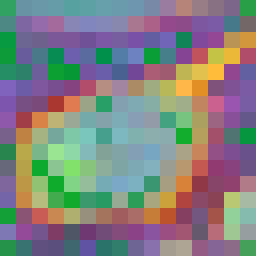} &
    \includegraphics[width=0.10\linewidth]{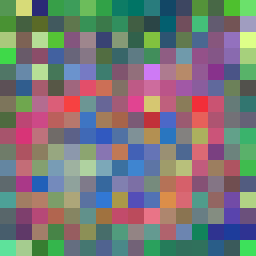} &
    \includegraphics[width=0.10\linewidth]{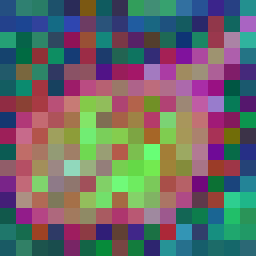} &
    \includegraphics[width=0.10\linewidth]{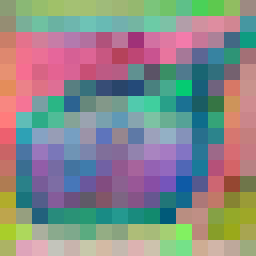} &
    \includegraphics[width=0.10\linewidth]{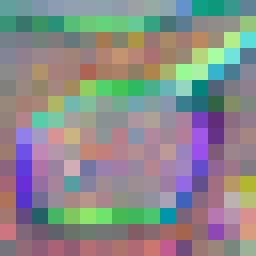} \\[2pt]
    \includegraphics[width=0.10\linewidth]{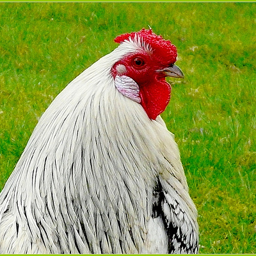} &
    \includegraphics[width=0.10\linewidth]{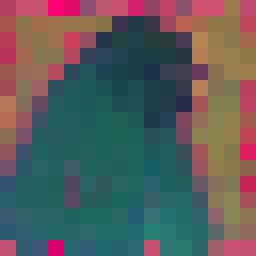} &
    \includegraphics[width=0.10\linewidth]{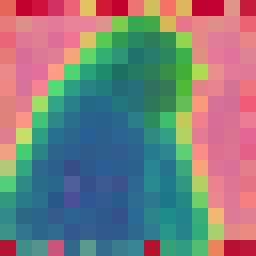} &
    \includegraphics[width=0.10\linewidth]{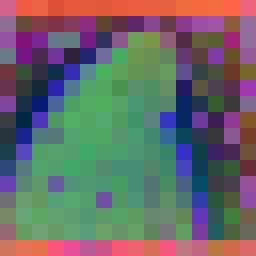} &
    \includegraphics[width=0.10\linewidth]{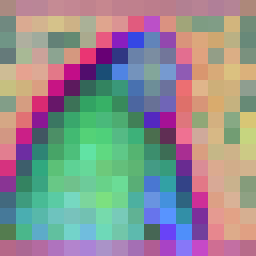} &
    \includegraphics[width=0.10\linewidth]{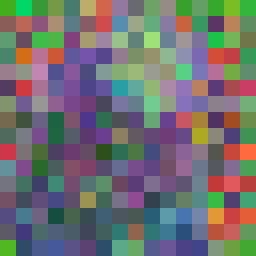} &
    \includegraphics[width=0.10\linewidth]{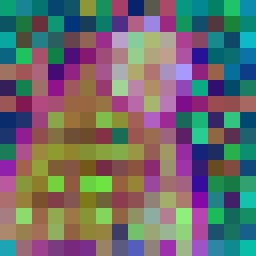} &
    \includegraphics[width=0.10\linewidth]{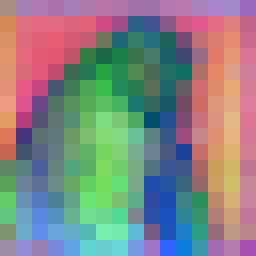} &
    \includegraphics[width=0.10\linewidth]{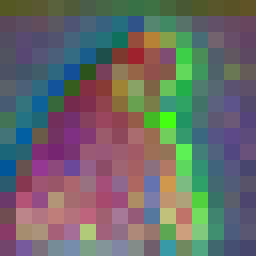} \\[2pt]
    \includegraphics[width=0.10\linewidth]{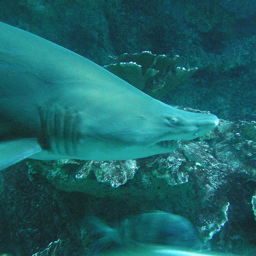} &
    \includegraphics[width=0.10\linewidth]{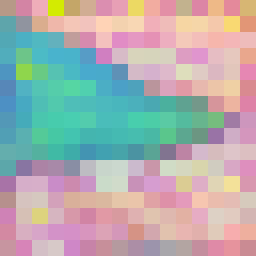} &
    \includegraphics[width=0.10\linewidth]{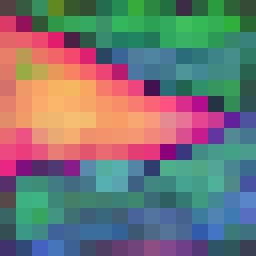} &
    \includegraphics[width=0.10\linewidth]{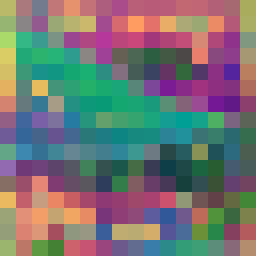} &
    \includegraphics[width=0.10\linewidth]{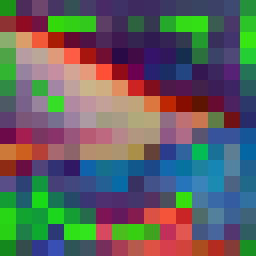} &
    \includegraphics[width=0.10\linewidth]{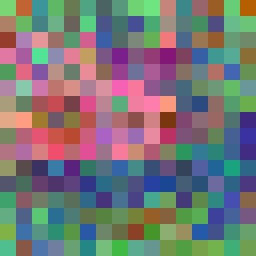} &
    \includegraphics[width=0.10\linewidth]{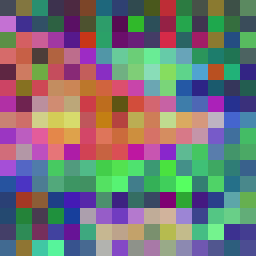} &
    \includegraphics[width=0.10\linewidth]{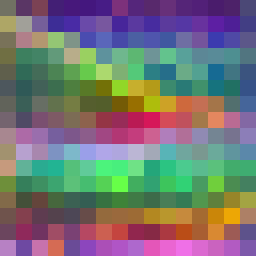} &
    \includegraphics[width=0.10\linewidth]{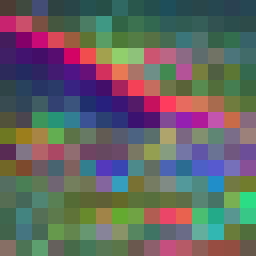} \\[2pt]
    \includegraphics[width=0.10\linewidth]{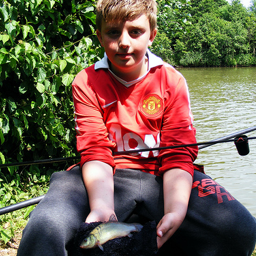} &
    \includegraphics[width=0.10\linewidth]{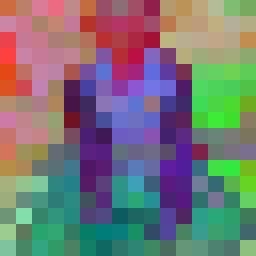} &
    \includegraphics[width=0.10\linewidth]{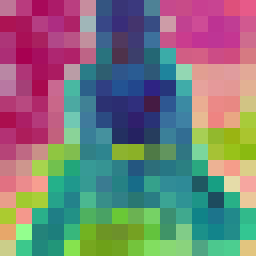} &
    \includegraphics[width=0.10\linewidth]{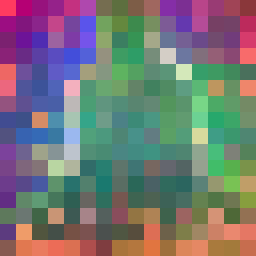} &
    \includegraphics[width=0.10\linewidth]{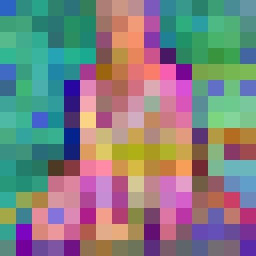} &
    \includegraphics[width=0.10\linewidth]{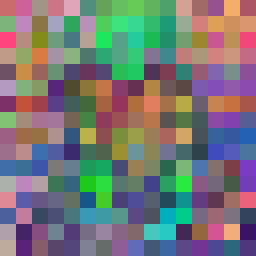} &
    \includegraphics[width=0.10\linewidth]{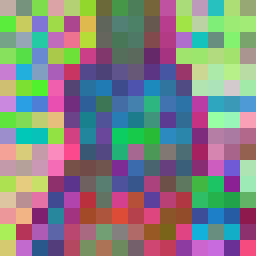} &
    \includegraphics[width=0.10\linewidth]{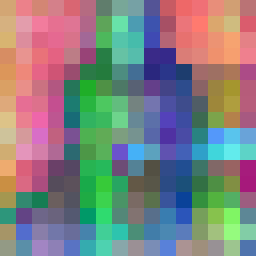} &
    \includegraphics[width=0.10\linewidth]{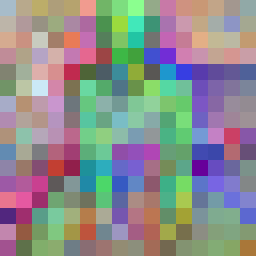} \\[2pt]
    \multicolumn{1}{c}{} &
    \multicolumn{2}{c}{\color{teal!80}\textbf{DINOv2}} &
    \multicolumn{2}{c}{\color{purple!60}\textbf{MOCOv3}} &
    \multicolumn{2}{c}{\color{mauve}\textbf{SigLIPv2}} &
    \multicolumn{2}{c}{\color{slate}\textbf{MAE}} \\
\end{tabular}
\caption{Qualitative comparison of feature visualizations. For each image, we show PCA visualization of \textcolor{teal!80}{DINOv2}, \textcolor{purple!60}{MOCOv3}, \textcolor{mauve}{SigLIPv2} and \textcolor{slate}{MAE} features. For each feature, we visualize PCA vs \texttt{CoReDi}'s learned projection.}
\label{fig:pca_vis_appendix}
\end{figure}

\end{document}